\documentclass[11pt, a4paper, logo, copyright, nonumbering]{cls/Sci}

\usepackage[numbers, sort&compress, square]{natbib}
\usepackage{url}
\usepackage{amsmath}
\usepackage[table]{xcolor}
\usepackage{multirow}
\usepackage{makecell}
\usepackage{booktabs}
\usepackage{amsfonts}
\usepackage{nicefrac}
\usepackage{graphicx}
\usepackage{enumitem}
\usepackage{pifont}
\usepackage{pdflscape}
\usepackage{array}
\usepackage{caption}
\usepackage{subcaption}
\usepackage{wrapfig}
\usepackage{float}
\usepackage{placeins}
\usepackage{listings}
\usepackage{fvextra}
\usepackage{adjustbox}
\usepackage{fontawesome5}
\usepackage{textcomp}
\usepackage{etoolbox}
\usepackage[most]{tcolorbox}

\lstdefinestyle{python}{
  language=Python,
  basicstyle=\ttfamily\small,
  keywordstyle=\color{blue}\bfseries,
  stringstyle=\color{red!70!black},
  commentstyle=\color{green!50!black}\itshape,
  showstringspaces=false,
  breaklines=true,
  frame=single,
  framerule=0.4pt,
  rulecolor=\color{black!30},
  backgroundcolor=\color{black!3},
  numbers=left,
  numberstyle=\tiny\color{black!40},
  numbersep=6pt,
  tabsize=4,
  literate=
    {->}{{\textrightarrow}}2
    {None}{{\textbf{None}}}1
    {True}{{\textbf{True}}}1
    {False}{{\textbf{False}}}1,
}
\title{Agentic Time Machine as an Infrastructure for Future-Event Forecasting}

\author{
{\normalfont\mdseries
Jingyi Chai\textsuperscript{1,*},
Bingyang Zheng\textsuperscript{1,*},
Xiangrui Liu\textsuperscript{1},
Hao Lu\textsuperscript{1},
Zihang Zhou\textsuperscript{1},
Tianchen Wang\textsuperscript{1},
Kemeng Zhang\textsuperscript{1},
Siheng Chen\textsuperscript{1,\ensuremath{\dagger}}
}
\\
{\normalfont\bfseries
\textsuperscript{1}Shanghai Jiao Tong University
}
\\
{\normalfont\mdseries
\textsuperscript{*}Equal contribution.
\quad
\textsuperscript{\ensuremath{\dagger}}Corresponding author: sihengc@sjtu.edu.cn
}
}

\begin{document}

\begin{abstract}
Forecasting future events is a critical challenge for large language model (LLM) agents, spanning domains from elections and monetary policy to financial markets. However, evaluating progress on this task presents a fundamental trade-off between efficiency and environment fidelity. While live evaluation benchmarks suffer from an inherently slow feedback loop, existing retrospective replays typically restrict agents to static, pre-frozen databases that sacrifice the environmental realism of actual deployments. To tackle this issue, we introduce Agentic Time Machine (TM), an infrastructure that approximately reconstructs the web state at any chosen past time by filtering post-cutoff content. Leveraging this evaluation infrastructure, we further propose a planner-solver-aggregator multi-agent framework that breaks each question into diverse analytical angles, gathers evidence in parallel, and combines the results into a single forecast. Experiments show that offline scores under TM correlate strongly with live FutureX scores, validating that TM offers a fast and reliable sandbox for forecasting-agent evaluation. On FutureX-Past and Polymarket evaluated under TM, our framework achieves the highest score among strong closed-book, tool-augmented, and self-consistency baselines. On the official FutureX live leaderboard, our system achieves the best average rank over four consecutive weeks, including \textbf{1st place} in May Week 1. As of June 17, \textbf{it also ranks 1st on FutureX's official eight-week overall leaderboard.}
\end{abstract}

\maketitle
\begin{figure}[h]
    \centering
    \includegraphics[width=1\linewidth]{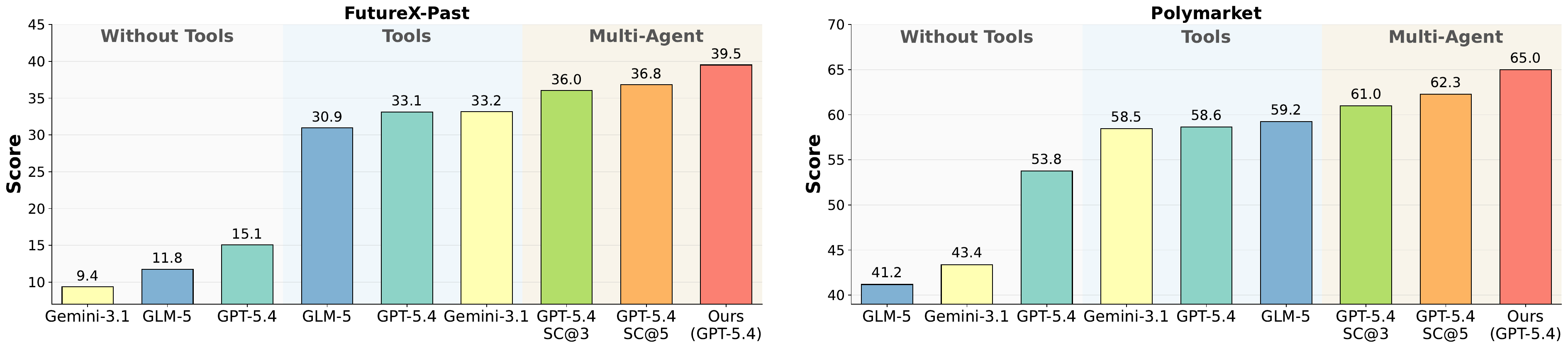}
    \caption{Main offline forecasting results on two locally constructed subsets. Both subsets are replayed under Agentic Time Machine to block post-cutoff answer leakage. Bars compare representative closed-book models, tool-using agents, and multi-agent variants.}
    \label{fig:main_bar}
\end{figure}

\section{Introduction}
\label{sec:intro}

Forecasting real-world events such as election outcomes, monetary-policy decisions, sports results, and asset price movements remains a critical challenge for large language model (LLM) agents~\cite{social_predictor,approaching}. Unlike deep information retrieval tasks~\cite{browsecomp}, the correct answer for a forecasting problem does not yet exist in any online source. The agent should collect useful evidence from many noisy sources, account for uncertainty, and reason about the potential outcomes. Recent live benchmarks such as FutureX~\cite{futurex}, Prophet Arena~\cite{ProphetArena}, and ForecastBench~\cite{forecastbench} facilitate evaluation of prediction capacity by releasing fresh questions periodically and scoring predictions once the real-world events resolve.

\begin{figure}[h]
    \centering
    \includegraphics[width=0.5\linewidth]{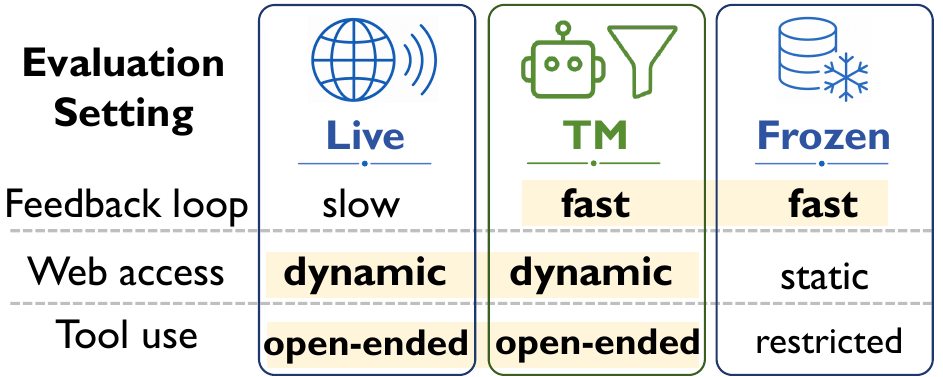}
    \caption{Comparison of evaluation settings for forecasting agents. Our Agentic Time Machine (TM) uniquely combines a fast feedback loop with dynamic, open-ended web interactions.}
    \label{fig:evaluation_modes}
\end{figure}

However, pushing the boundaries of real-world forecasting tasks presents a fundamental trade-off between evaluation efficiency and environment fidelity. 
As illustrated in Figure~\ref{fig:evaluation_modes}, there are two paradigms of existing evaluation benchmarks denoted as Live and Frozen. On one hand, \textbf{Live} evaluation environments~\cite{futurex,ProphetArena,forecastbench,foresightarena} offer dynamic web access and open-ended tool use, yet they are constrained by an inherently slow feedback loop, where researchers must wait days or weeks for real-world event resolution. On the other hand, to accelerate development cycles, alternative benchmarks explore a \textbf{Frozen} replay paradigm~\cite{chandak2025scaling,ye2024mirai,futuresim} or closed-book inference~\cite{dai2025dailyoracle,tan2025inferring}. Although these methods provide a fast feedback loop, they restrict the agent to retrieving information from a frozen database, fundamentally sacrificing the environmental realism of actual deployments.

To address this challenge, we propose \emph{Agentic Time Machine} (TM), an LLM-based evaluation infrastructure that achieves a fast feedback loop while preserving dynamic web access. By actively filtering post-cutoff or answer-revealing content from web tools, TM effectively simulates the open web as it stood at any chosen historical prediction point, preventing answer leakage~\cite{web_leakage} during retrospective replay. Moreover, to improve forecasting ability, we build a planner-solver-aggregator multi-agent system that decomposes each question into a small set of analytical angles, gathers evidence along those different paths in parallel, and combines these findings into a single final forecast.

We validate both components through extensive empirical evaluation. We first compare agent performance with and without TM across five LLM backbones, where applying TM substantially lowers scores on prediction tasks and shows that our infrastructure effectively reduces answer leakage. Importantly, we find that offline scores achieved under the Time Machine correlate strongly with live leaderboard performance, establishing our infrastructure as a reliable proxy for real-world prediction testing. Leveraging TM, our forecasting system achieves the highest overall scores on two offline prediction datasets. On the public FutureX live leaderboard, the same system, listed as FutureGPTv1, ranks among the leading submissions across four consecutive weekly rounds. As of June 17, 2026, it also ranks first on FutureX's official eight-week overall leaderboard. Together, these results suggest that fast and leakage-free local replay is a useful evaluation infrastructure, and planner-solver-aggregator multi-agent decomposition improves forecasting under both offline and live settings.

In summary, our contributions are as follows:
\begin{itemize}
\setlength{\itemsep}{1pt}
\item \textbf{Agentic Time Machine}, an LLM-based evaluation infrastructure that simulates the web as it stood at a chosen point in time, enabling fast and reproducible local replay of forecasting tasks without exposing the resolved answer to the forecasting agent (Section~\ref{sec:bench}).
\item \textbf{A planner-solver-aggregator multi-agent forecasting system} in which a planner decomposes each question into diverse analytical angles, parallel solvers gather evidence, and an aggregator reconciles the resulting views, with no training and no extra tools beyond a single-agent baseline (Section~\ref{sec:method}).
\item \textbf{Comprehensive empirical evidence} showing that Time Machine provides a fast and reliable infrastructure for local forecasting evaluation, and that our planner-solver-aggregator framework improves forecasting performance across offline and live settings (Section~\ref{sec:exp}).
\end{itemize}

\section{Related Work}
\subsection{LLM Prediction Benchmarks}

\textbf{Continuously updated forecasting benchmarks.}
A growing line of work evaluates LLMs on upcoming future events rather than retrospective question answering, establishing a \textbf{Live} evaluation paradigm. ForecastBench~\cite{forecastbench} is a dynamic benchmark for forecasting under genuine uncertainty, while FutureX~\cite{futurex} further emphasizes the agentic setting by evaluating tool-using LLM agents on diverse live prediction tasks. Prophet Arena~\cite{ProphetArena} studies predictive intelligence in a similarly live setup, and DailyOracle~\cite{dai2025dailyoracle} continuously generates future-oriented evaluation instances from daily news to track temporal generalization. Although these live benchmarks offer authentic dynamic web access and open-ended tool use, they are constrained by an inherently slow feedback loop, as improving a system requires waiting days or weeks for real-world event resolution.

\textbf{Retrospective replay and pastcasting benchmarks.}
To shorten this loop, another line of work implements a \textbf{Frozen} replay paradigm by replaying historical forecasting questions under temporally constrained conditions. Bench to the Future~\cite{pastcasting_bench} converts resolved questions into an offline benchmark with pre-collected evidence, and FutureSim~\cite{futuresim} replays world events chronologically so that agents observe information as it unfolds over time. Similarly, MIRAI~\cite{ye2024mirai} evaluates event forecasting in a temporally grounded setting. While these retrospective benchmarks enable a fast feedback loop, they typically rely on frozen web snapshots or simulated replay environments that prevent agents from dynamically interacting with the vast real-time web information, thereby lacking dynamic web access and enforcing restricted tool use. Moreover, despite date-restricted search, naive retrospective evaluation remains highly vulnerable to post-hoc answer leakage through updated web snippets and pages with rolling content~\cite{web_leakage}.
Unlike these approaches, our Agentic Time Machine (TM) retains the efficiency of local replay while preserving open-ended, dynamic web interaction and explicitly preventing answer leakage.

\subsection{Multi-Agent Forecasting Systems}
Recent work has explored multiple agents or multiple sampled reasoning paths. A simple form is self-consistency~\cite{sc}, which aggregates independently sampled solutions but mainly diversifies answers rather than the evidence-gathering process itself. More forecasting-specific systems go further. AIA Forecaster~\cite{aia} combines agentic search with a supervisor agent that reconciles forecasts and applies calibration, while MAEPS~\cite{maeps} simulates human expert-team collaboration through specialized agents that gather evidence from different professional dimensions and reconcile temporally conflicting information. MiroFlow~\cite{miroflow} provides broader evidence that careful orchestration can strengthen complex research-style agent systems. While these prior works highlight the value of multi-agent coordination, they often rely on centralized search without explicitly diversifying retrieval perspectives.
In contrast, our planner-solver-aggregator framework decomposes each question into a small set of analytical angles, drives parallel solvers along distinct evidence paths, and aggregates their reports into one final forecast.

\begin{figure*}[t]
    \centering
    \includegraphics[width=1\linewidth]{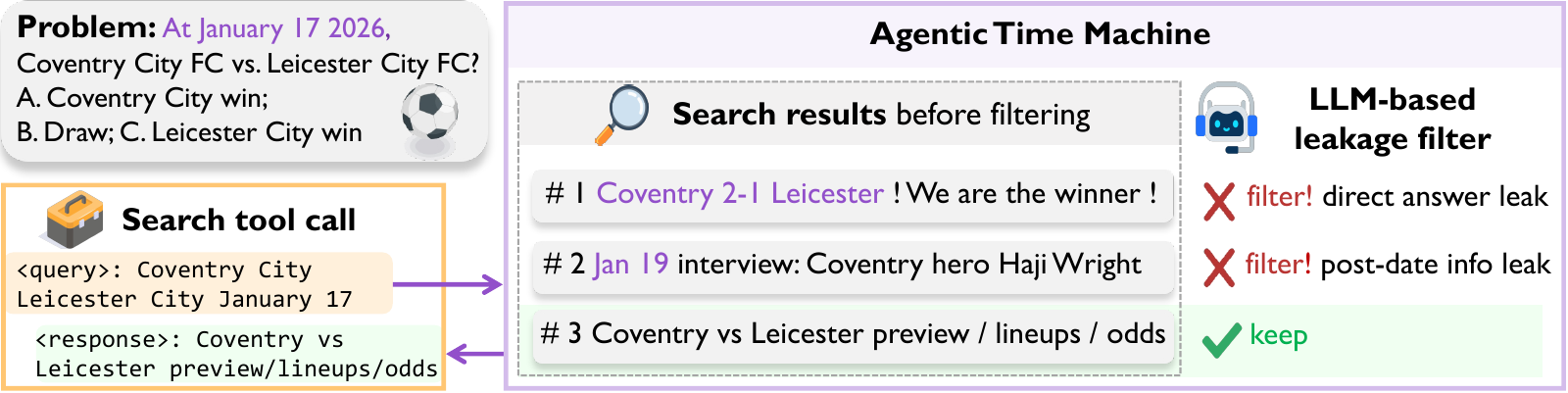}
    \caption{Agentic Time Machine filters web results to approximate the information environment available at a question's cutoff time. In this example, snippets that directly reveal the answer or contain post-cutoff information are removed, leaving only safe pre-cutoff evidence for forecasting.}
    \label{fig:atm}
\end{figure*}
\section{Offline Prediction Evaluation with Agentic Time Machine} 
\label{sec:bench}
To enable fast iteration without waiting for real-world resolution, we build a controlled local test bed on top of the FutureX historical archive. The dataset itself is not a contribution of this paper; the contribution is the Time Machine filter that makes local replay faithful to the agentic setting that the live benchmark targets.

\subsection{Problem Formulation}
\label{sec:problem}

A future-event forecasting instance is a tuple $(q, t, a^\star)$, where $q$ is a question about a real-world event, $t$ is the question's \emph{cutoff time}, and $a^\star$ is the ground-truth answer that becomes publicly known only after the event resolves at some time $t_{\text{res}} > t$. Given $q$ and an information environment $\mathcal{E}$ (the body of web evidence the agent can reach via web tools), an LLM agent $\pi$ outputs a prediction $\hat a \sim \pi(q, \mathcal{E})$, scored against $a^\star$ by a metric $r(\hat a, a^\star) \in [0,1]$. Unlike standard QA, the resolved outcome is not yet publicly known at time $t$; the forecasting objective is therefore defined with respect to the \emph{pre-cutoff} environment $\mathcal{E}_t$ that would have been visible at time $t$:
\begin{equation}
\max_\pi \; \mathbb{E}_{(q, t, a^\star) \sim \mathcal{D},\; \hat a \sim \pi(q, \mathcal{E}_t)} \big[\, r(\hat a, a^\star) \,\big].
\end{equation}

Since $\mathcal{E}_t$ is not directly accessible at evaluation time $t' > t$, local evaluation must substitute some approximation $\hat{\mathcal{E}} \approx \mathcal{E}_t$, and this choice shapes what is actually measured:
\begin{itemize}
\setlength{\itemsep}{1pt}
\item \textbf{Live web} $\hat{\mathcal{E}} = \mathcal{E}_{t'>t}$: contains content created after $t$ that may reveal $a^\star$; the task collapses from forecasting into retrieval.
\item \textbf{Frozen snapshot} $\hat{\mathcal{E}} = \mathcal{E}_{\text{frozen}} \subset \mathcal{E}_t$: a small pre-curated subset of $\mathcal{E}_t$ avoids leakage but its coverage and interaction pattern no longer match the live web that agents are deployed against.
\item \textbf{Time Machine} $\hat{\mathcal{E}} = \mathcal{E}_{\text{TM}} \approx \mathcal{E}_t$: apply an LLM-based leakage filter on the live web to remove content dated after $t$ or that exposes $a^\star$, so that the agent's effective environment approximates $\mathcal{E}_t$ while preserving open-ended tool interaction.
\end{itemize}

\subsection{Agentic Time Machine}
\label{sec:tm}

Agentic Time Machine instantiates the environment $\mathcal{E}_{\text{TM}}$ on top of the live web. Ideally, we would retain only those items in $\mathcal{E}_{t'>t}$ that were already retrievable with the same content at time $t$:
\begin{equation}\label{eq:tm-idea}
  \mathcal{E}_{\text{TM}}^{\star}=\{x \in \mathcal{E}_{t'>t} : x \text{ retrievable at } t\,\} \approx \mathcal{E}_t.
\end{equation}
Directly determining whether an item was retrievable at $t$ is impractical, so we approximate Equation~\ref{eq:tm-idea} with a binary leakage filter $\phi(x \mid q, t, a^\star)$, conditioned on the question $q$, its cutoff $t$, and the resolved answer $a^\star$. Rather than deciding whether $x$ was literally retrievable at $t$, the filter looks for two tractable signs of leakage. A \emph{direct answer leak} fires when the item states or strongly implies $a^\star$. A \emph{post-date info leak} fires when the item carries a time signal later than $t$, such as a publish date, a last-update date, or an in-page event date. An item is kept only if neither sign is present.

We assume the agent reaches the web only through a fixed set of web tools, in our implementation a web-search tool and a page-content tool, each with its own tool-specific prompt for the LLM filter (see Appendix~\ref{app:tm-mechanics} for more details). Each tool call returns items $\mathcal{X} \subseteq \mathcal{E}_{t'>t}$, of which the agent actually sees
\begin{equation}\label{eq:tm-filter}
  \mathcal{X}^{\text{TM}} = \{\, x \in \mathcal{X} : \phi(x \mid q, t, a^\star) = \textsc{keep}\,\}.
\end{equation}
Aggregating Equation~\ref{eq:tm-filter} over every tool call in a trajectory gives the agent's effective environment $\hat{\mathcal{E}} = \mathcal{E}_{\text{TM}} = \bigcup_{\text{calls}} \mathcal{X}^{\text{TM}}$, an approximation but not in general an equality with the ideal $\mathcal{E}_{\text{TM}}^{\star}$: a pre-cutoff article that strongly hints at the resolved answer can be over-blocked, and a post-cutoff page that happens to expose neither leakage sign can slip through. We characterize both error modes empirically in Appendix~\ref{app:tm-overfilter}.

Concretely, Agentic Time Machine wraps every web-tool call before the result reaches the forecasting agent. For search, the LLM filter checks each returned title and snippet. For page visits, it checks the retrieved page content. In both cases, the filter asks whether the content is outside the allowed time range, or whether it directly reveals the answer to the question. Items that fail either check are removed, and only the remaining items are passed to the agent. A typical case is sketched in Figure~\ref{fig:atm} on a Coventry-vs-Leicester forecasting question dated January~17, 2026. The filter drops a snippet titled "Coventry 2-1 Leicester" as a direct answer leak and drops a January~19 post-match interview as a post-date info leak.

\textbf{Why an LLM filter instead of a rule-based date cutoff.} A simpler design is to filter each retrieved URL by a single date stamp. We find this insufficient because a single date stamp cannot describe a page whose content evolves over time. Search engines attribute one heuristic date to each URL (some combination of creation, first-indexing, and in-page dates), which is not the same as the date on which the answer-bearing content was actually written. Many forecasting-relevant pages have a stable URL and an old indexed date but rolling body content. The Yahoo Finance historical-prices page for a stock such as AAPL is one such page: its URL and indexed date are stable, yet its in-snippet OHLC table extends to whichever date the reader queries, including the very date asked about by a forecasting question. So a date filter passes the URL through, but the snippet still reveals the high of day. A pre-cutoff news article later edited to add the resolved outcome behaves similarly. 

\section{Planner-Solver-Aggregator Multi-Agent Framework} 
\label{sec:method}
\begin{figure}[t]
    \centering
    \includegraphics[width=0.7\linewidth]{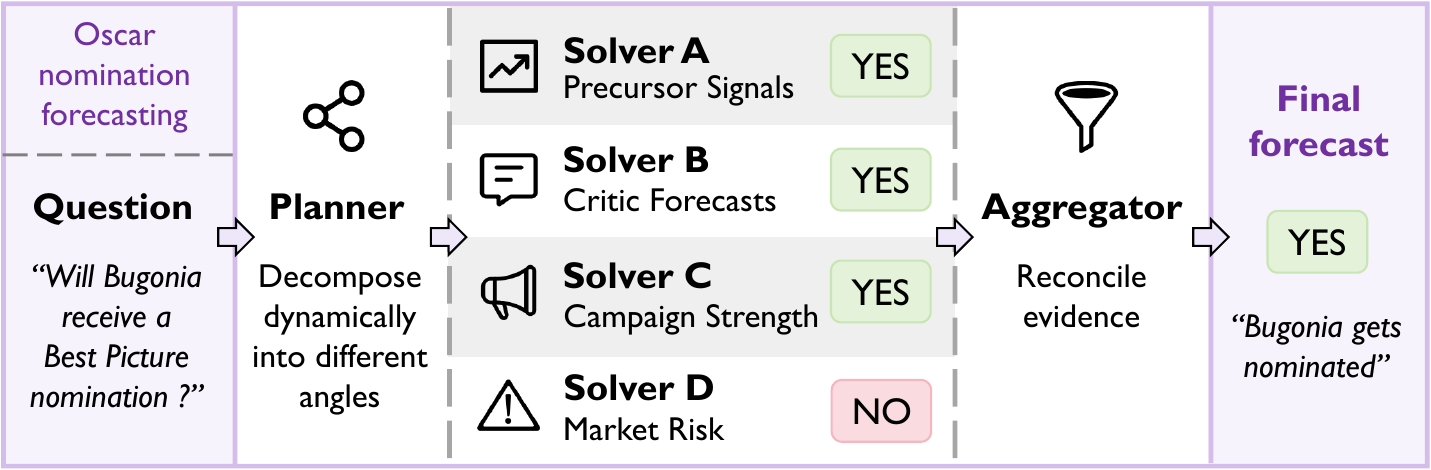}
    \caption{Overview of our planner-solver-aggregator multi-agent forecasting framework, illustrated on a Best Picture nomination question.}
    \label{fig:framework}
\end{figure}
Forecasting an unresolved event is rarely a single-path reasoning problem. The same question can usually be approached from several legitimate angles, such as historical base rates and analogies, structural causal drivers, recent leading indicators, and contrarian risks. Each angle exposes a different slice of the relevant evidence. A capable forecaster typically does not commit to one angle in advance, but explores several in parallel and then reconciles what they imply.

Our framework follows this structure, as shown in Figure~\ref{fig:framework}. A planner decides which angles to explore for the question at hand. Several solver agents cover those angles in parallel, each with its own search and reasoning. A final aggregator reconciles their reports into a single grounded answer. All three roles are equipped with web tools.

\textbf{Planner.} Given the prediction question, the planner performs a few exploratory searches and produces a JSON plan with (a) a short task understanding, (b) the dominant causal factors, (c) historical patterns it has confirmed, and (d) 2--4 solver profiles. Each role profile defines a unique angle, a factor path the solver should follow, the search focus, and what the solver must deliver. The planner also writes aggregation advice.

\textbf{Solvers.} The 2--4 solvers run in parallel on separate threads. Each receives only the planner's understanding, shared key factors, and its own spec. The solver is instructed to use tools actively, ground its claims in evidence rather than priors, and produce a structured JSON report. Because each solver follows a distinct angle, the system can cover more relevant evidence without forcing one agent to pursue every possible line of reasoning.

\textbf{Aggregator.} The aggregator receives the planner output and all solver reports, identifies points of agreement and conflict, runs additional targeted searches when needed, and writes a free-form synthesis; the final line matches the question's required answer format. 

Our framework adds no training, no fine-tuning, and no extra tools beyond what a single-agent baseline already has. Its main design choices are (i) a planner that first researches the prediction problem and assigns plausible forecasting angles, (ii) parallel angle-conditioned solvers that gather evidence from different paths, and (iii) an aggregator that reconciles individual views.

\section{Experiments}
\label{sec:exp}

\subsection{Setup}
\label{sec:exp-setup}

\textbf{Datasets.} For offline evaluation, we use two historical question pools replayed via Agentic Time Machine. The first pool is a 61-question subset sampled from the FutureX historical archive~\cite{futurex}, which we refer to as FutureX-Past. It covers all four official difficulty levels: 17 at L1 (few options), 17 at L2 (multi-option with non-trivial domain context), 9 at L3 (compositional questions that chain several intermediate facts), and 18 at L4 (quantitative or open-ended). The second pool comprises 80 L2-style multi-option questions sampled from Polymarket, a real-money prediction market covering politics, finance, sports, and pop culture. We additionally evaluate our framework on the live, dynamic FutureX leaderboard, where real-world questions are still unresolved at submission time, thus Time Machine is not required for the live setting. Dataset construction details for both pools are in Appendix~\ref{app:data}, and a deeper analysis of our Time Machine is in Appendix~\ref{app:tm}.

\textbf{Baselines.} We evaluate three categories of methods, all using a temperature of 1.0: (1) Models, six closed-book LLMs without tool access; (2) Agents, the same LLMs equipped with two Time Machine-wrapped tools (web search and page fetching) via a ReAct-style loop \cite{react}; and (3) Multi-Agent, including self-consistency (SC$@N$ for $N\in \{3,5\}$), which run the agent forecaster N times and majority-vote the final answer~\cite{sc}, as well as our planner-solver-aggregator framework, all of which utilize a GPT-5.4-medium~\cite{gpt54} backbone and two web tools. For most configurations, we report the average score of three independent runs. However, for SC$@N$, we report a single ensemble run, as it already incorporates $N$ forecasts. See web tool details in Appendix~\ref{app:web_tools}.

\textbf{Metric.} We follow the official FutureX scoring protocol~\cite{futurex}. For L1 and L2, we report set-level F1, which reduces to exact match for single-option questions. For L3 and L4, an LLM judge evaluates predictions against resolved answers using three labels (Yes/Partial Correct/No); numeric L4 items additionally receive a graded distance score, $\max(0, 1 - ((\hat{y}-y))^2)$, to reward near-miss predictions. All per-question scores lie in $[0,1]$. For \textsc{FutureX-Past}, we report both per-level means $(m_\ell)$ and the weighted aggregate score $0.1m_1 + 0.2m_2 + 0.3m_3 + 0.4m_4$, which emphasizes harder levels. For Polymarket, we report mean set-level F1 across the 80 L2-style items. All scores are scaled by $100$ for readability.

\subsection{Agentic Time Machine Validation}
\label{sec:exp-timemachine}

\begin{figure*}[t]
  \centering
  \begin{minipage}[t]{0.48\textwidth}
    \vspace{0pt}
    \centering
    \begin{small}
    \renewcommand{\arraystretch}{1.15}
      \setlength{\tabcolsep}{2.2pt}
      \begin{tabular}{l|ccc}
        \toprule
        Model & \makecell{w/o TM} & \makecell{w/ TM} & $\Delta$ \\
        \midrule
        Kimi-K2.5~\cite{kimik25} & 60.64 & 27.13 & -33.51 \\
        GLM-5~\cite{glm5} & 67.45 & 30.94 & -36.51 \\
        \makecell[l]{DeepSeek-V4-Flash~\cite{deepseekv4flash}} & 69.11 & 35.02 & -34.09 \\
        GPT-5.4-medium~\cite{gpt54} & 64.39 & 33.13 & -31.26 \\
        \makecell[l]{Claude-4.6-Sonnet~\cite{claudeSonnet46}} & 47.81 & 22.80 & -25.01 \\
        \bottomrule
      \end{tabular}
    \end{small}
    \captionof{table}{Effects of Time Machine (TM) on the same 61-question set with tool use enabled. All results are averaged over three independent runs. For the w/o TM setting, the web tools have unrestricted access to the open web. All scores are scaled by $100$ for readability.}
    \label{tab:tm}
  \end{minipage}
  \hfill
  \begin{minipage}[t]{0.48\textwidth}
    \vspace{0pt}
    \centering
    \includegraphics[width=0.95\linewidth]{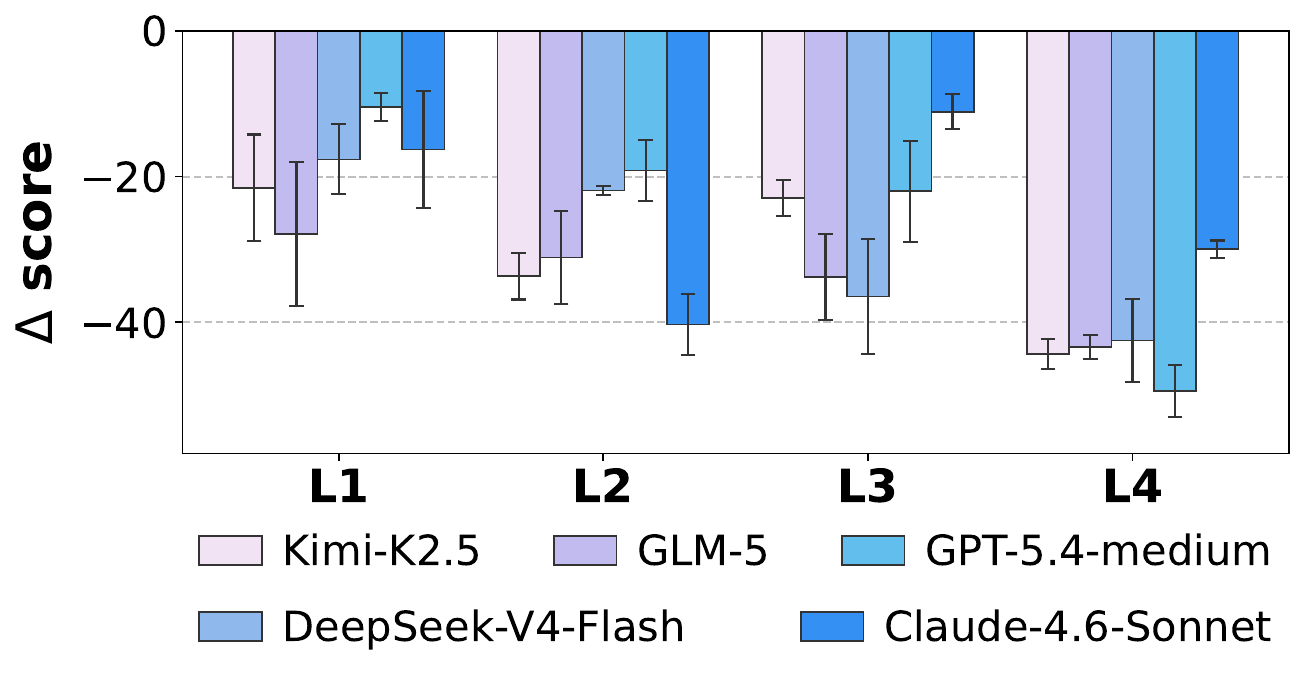}
    \captionof{figure}{Per-level $\Delta$ score (w/ TM $-$ w/o TM, in score points on the $0$--$100$ scale) across five backbones, with $\pm 1$ standard deviation error bars over 3 runs per cell. Bars are grouped by FutureX difficulty level and each color denotes one model.}
    \label{fig:tm-per-level}
  \end{minipage}
\end{figure*}

Table~\ref{tab:tm} shows how the Agentic Time Machine affects performance across five different backbones. Without the Time Machine, most models achieve a weighted score between $60$ and $69$. This high performance occurs because agents can often find the final answers directly on the live web. However, enabling the Time Machine consistently reduces these scores by $25$ to $37$ points across different model families and sizes. The consistency of this drop suggests that the Time Machine is effectively blocking a common source of information leakage. Based on these results, we apply the Agentic Time Machine to all subsequent offline tool-using experiments.

\textbf{Where the drop comes from.} By breaking the aggregated drop down by question level in Figure~\ref{fig:tm-per-level}, we find that (1) Level 4 is the most heavily affected by information leakage, as scores for all backbones fall to a low range between $12$ and $29$ with the Time Machine (TM) enabled. This sharp decrease occurs because Level 4 consists of open-ended and quantitative questions rather than multiple-choice items, making performance highly dependent on advanced forecasting and reasoning skills once direct web answers are blocked. We also find that (2) Level 1 questions show a much smaller change, and their scores stay within a tight range of $68$ to $72$. This minor drop of only $10$ to $28$ points occurs because these few-option questions inherently offer a high random guessing baseline, allowing models to maintain solid scores by leveraging historical baselines, public polls, and prior news to rule out incorrect choices even when future web data is completely hidden. Finally, we find that (3) beyond these two extremes, intermediate levels reveal model-specific patterns, where Claude-4.6-Sonnet drops most on Level 2 while its low baseline scores on harder levels limit further decreases, whereas GPT-5.4-medium shows the largest Level 4 drop, indicating its numerical forecasting was heavily driven by open-web leaks rather than robust reasoning. The means and standard deviations are reported in Appendix~\ref{app:tm-perlevel}, and two Level-4 case studies are in Appendix~\ref{app:tm-cases}.

\textbf{Why w/o TM scores are not near 100.} Even with unrestricted web access, scores without TM are between $60$ and $69$ for most models instead of reaching the perfect score of $100$. By analyzing the failed cases, we find that three distinct factors limit this performance. Specifically, we observe that (1) some source webpages have been deleted, placed behind paywalls, or removed from search engine indexes over time, meaning this information is no longer retrievable on the live web. We also find that (2) even when the relevant pages still exist, the search queries generated by the agent or the ranking algorithms of the search engine sometimes fail to find them. Finally, we find that (3) even if the agent successfully surfaces a useful page, it occasionally misinterprets the text or chooses not to trust the information. Consequently, both settings have an upper bound on performance, and current agents have substantial room for improvement in both environments. Therefore, the $25$ to $37$ point drop shown in Table~\ref{tab:tm} indicates that information leakage significantly inflates open-web performance.

\subsection{Offline Results with Agentic Time Machine}
\label{sec:exp-local}

\begin{table*}[t]
  \begin{center}
      \setlength{\tabcolsep}{7pt}
      \begin{tabular}{l|rrrrc|c}
        \toprule
        \textbf{Dataset} & \multicolumn{5}{c|}{\textbf{FutureX-Past}} &  \textbf{Polymarket}  \\
        \textbf{Metric}                                    & \textbf{L1}       & \textbf{L2}        & \textbf{L3}        & \textbf{L4}        & \textbf{Score} & \textbf{Score}       \\
        \midrule
        \rowcolor{blue!15} \multicolumn{7}{c}{\textbf{\textit{Models (closed-book, no tools)}}}                                                                                                  \\
        GPT-5.4-medium~\cite{gpt54}                            & 66.67          & 21.62          & 1.01           & 9.45           & 15.06   & 53.76        \\
        Claude-4.6-Sonnet~\cite{claudeSonnet46}                        & 62.75          & 15.65          & 0.00           & 0.99           & 9.78    & 47.97        \\
        Gemini-3.1-Pro~\cite{gemini31pro}                        & 54.90          & 12.83          & 0.00           & 3.33           & 9.36  & 43.38          \\
        Kimi-K2.5~\cite{kimik25}                         & 64.70          & 22.76          & 1.51           & 1.48           & 12.08    & 42.88       \\
        DeepSeek-V4-Flash~\cite{deepseekv4flash}                  & 54.90          & 14.09          & 4.69           & 1.70           & 10.39   & 34.79        \\
        GLM-5~\cite{glm5}                                    & 60.79          & 18.85          & 1.96           & 3.33           & 11.75  & 41.18         \\
        \midrule
        \rowcolor{blue!15} \multicolumn{7}{c}{\textbf{\textit{Agents (single-agent w/ tools, Time Machine on)}}}                                                                                  \\
        GPT-5.4-medium w/ tools                   & 69.98          & 66.75          & \underline{12.59}          & 22.52          & 33.13   & 58.63        \\
        Claude-4.6-Sonnet w/ tools                & 71.94          & 43.21          & 6.91           & 12.22          & 22.80    & 48.21       \\
        Gemini-3.1-Pro w/ tools                   & 68.63          & 62.18          & 7.90           & 28.72          & 33.16   & 58.45        \\
        Kimi-K2.5 w/ tools                        & 68.63          & 53.72          & 7.90           & 17.90          & 27.13   & 57.91        \\
        DeepSeek-V4-Flash w/ tools                & 68.63          & 64.00          & 12.42          & \underline{29.07} & 35.02   & 48.12        \\
        GLM-5 w/ tools                            & 68.17          & 58.24          & 8.39           & 24.90          & 30.94   & 59.24        \\
        \midrule
        \rowcolor{blue!15} \multicolumn{7}{c}{\textbf{\textit{Multi-Agent (Time Machine on)}}}                                                                        \\
        GPT-5.4-med-SC@3~\cite{sc} & 82.35 & \underline{77.89} & 8.89 & 23.90 & 36.04 & 60.99 \\
        GPT-5.4-med-SC@5~\cite{sc} &  \textbf{88.24} & \textbf{78.82} & 8.89 & 23.90 & \underline{36.81} & \underline{62.26}    \\
        \textbf{Ours} (GPT-5.4-medium) & \underline{84.31} & 71.72 & \textbf{14.07} & \textbf{31.28}          & \textbf{39.51} &  \textbf{64.99} \\
        \bottomrule
      \end{tabular}
  \end{center}
  \caption{Main results on FutureX-Past and Polymarket datasets under Time Machine. For FutureX-Past, we report the four official level scores and their weighted aggregate $0.1\,m_1 + 0.2\,m_2 + 0.3\,m_3 + 0.4\,m_4$. For Polymarket, the score is the unweighted mean set-level F1 over the 80 items. Non-ensemble rows are averaged over three independent runs, while SC@$N$ rows are reported from one ensemble run. The best value in each column is shown in \textbf{bold} while the second-best value is \underline{underlined}.}\label{tab:main}
\end{table*}

\begin{figure}[t]
    \centering
    \includegraphics[width=0.6\linewidth]{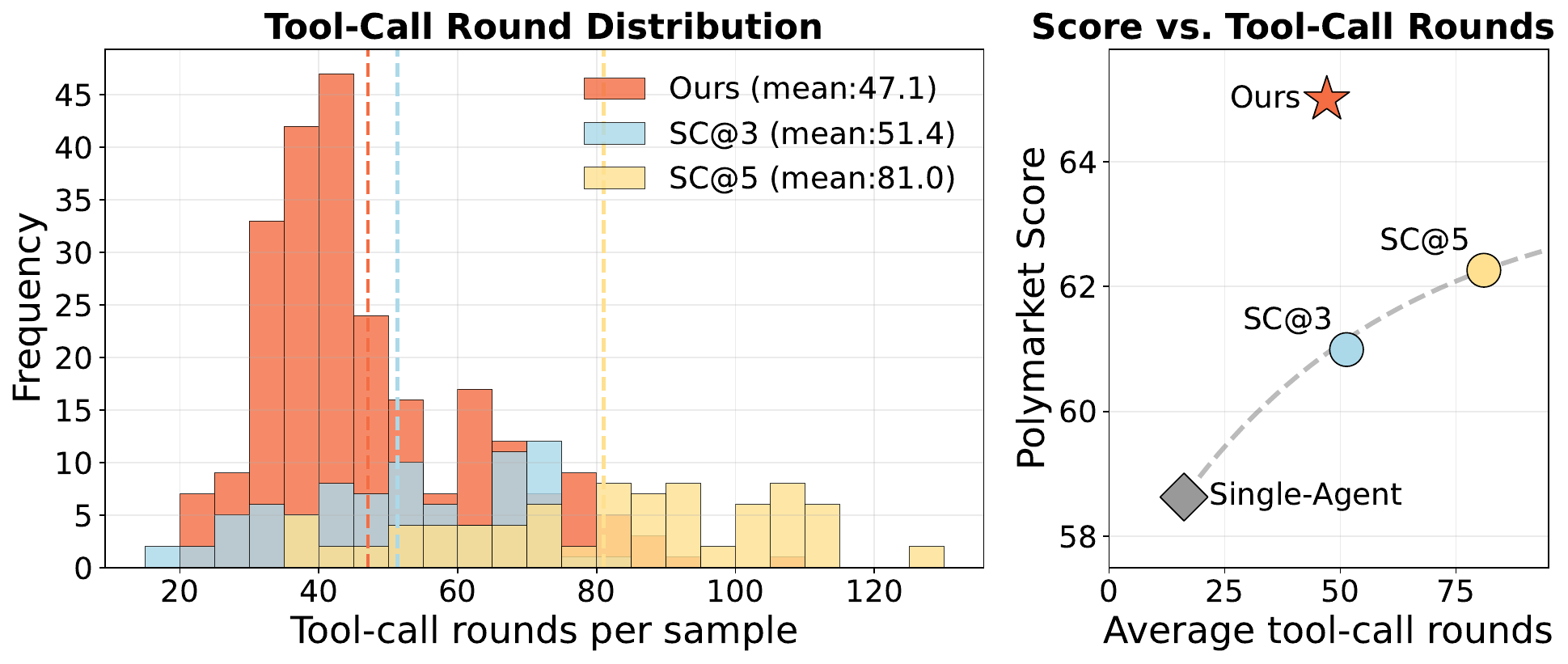}
    \caption{Tool-call efficiency of the multi-agent forecaster compared to self-consistency baselines on Polymarket. \textbf{Left}: per-question tool-call distribution across the 80-question pool; dashed lines mark per-method means (Ours $47.1$, SC@3 $51.4$, SC@5 $81.0$). \textbf{Right}: Polymarket score versus average tool-call rounds.}
    \label{fig:efficiency}
\end{figure}

\textbf{Our framework leads on both prediction datasets.} On FutureX-Past, our framework reaches a weighted score of $39.51$, outperforming the best self-consistency ensemble, SC@5, at $36.81$ and the strongest single-agent baseline, DeepSeek-V4-Flash with tools, at $35.02$. On Polymarket, our method also obtains the highest score of $64.99$, placing it ahead of SC@5 at $62.26$ and the top single-agent baseline, GLM-5 with tools, at $59.24$. These two datasets differ substantially, as FutureX-Past spans four difficulty levels across diverse forecasting domains, while Polymarket consists entirely of L2-style multi-option questions from real-money prediction markets. The consistent performance gains demonstrate that the planner-solver-aggregator design transfers effectively across different question types instead of overfitting a single benchmark. Furthermore, the detailed performance by level reveals that while self-consistency is competitive on L1 and L2 tasks, our framework establishes its primary lead on the harder L3 and L4 levels, which carry most of the weighted score.

\textbf{Web tools are the largest single step from closed-book models to forecasting agents.} Adding Time Machine-wrapped web tools improves the performance of every backbone on FutureX-Past by $13$ to $25$ weighted points over its closed-book counterpart, and boosts several backbones by more than $10$ points on Polymarket. Even the strongest closed-book model on FutureX-Past, GPT-5.4-medium at $15.06$, remains well below the weakest tool-equipped agent, Claude-4.6-Sonnet with tools at $22.80$. This performance gap persists even though the Time Machine suppresses post-cutoff and answer-revealing pages, which suggests that the improvement stems from gathering background evidence rather than raw answer lookup. Controlled web access is a fundamental requirement for strong forecasting agents, and the remaining performance gap highlights the need for better evidence collection and organization.

\begin{figure*}[t]
    \centering
    \begin{minipage}{0.48\linewidth}
        \centering
        \includegraphics[width=\linewidth]{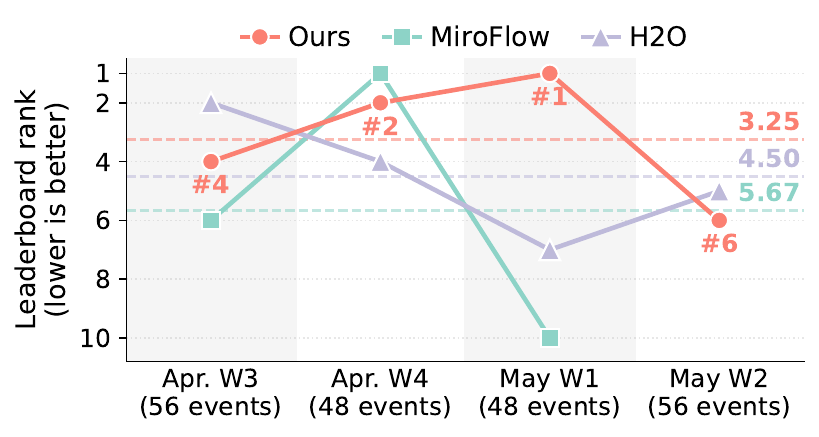}
        \caption{FutureX live leaderboard ranks across four consecutive weekly rounds, where a lower rank is better. Dashed horizontal lines show the average rank of each system over the weeks where it appears.}
        \label{fig:live-rank-trajectory}
    \end{minipage}
    \hfill
    \begin{minipage}{0.48\linewidth}
        \centering
        \includegraphics[width=\linewidth]{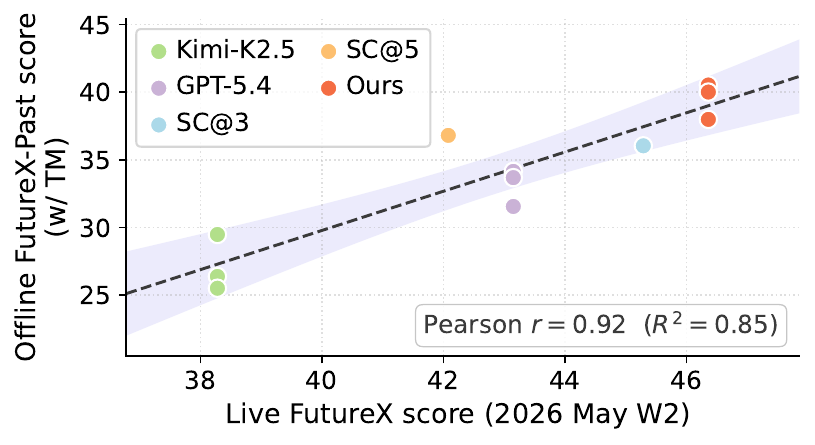}
        \caption{Consistency between offline scores with Time Machine and live FutureX scores from 2026~May~W2. Each point corresponds to one run of a method. The dashed line is a linear fit over 11 points.}
        \label{fig:live-offline-consistency}
    \end{minipage}
\end{figure*}

\textbf{Role decomposition beats majority voting on both score and efficiency.} To determine whether our gains come from the multi-agent architecture itself or simply from increased computation, we compare our framework against self-consistency baselines on Polymarket. Here, SC@3 and SC@5 reach scores of $60.99$ and $62.26$, which are both below our score of $64.99$. Figure~\ref{fig:efficiency} shows that our framework also uses fewer tool-call rounds on average, requiring $47.1$ rounds compared to $51.4$ for SC@3 and $81.0$ for SC@5. While SC@5 gains only $1.27$ points over SC@3 at the cost of $58\%$ more rounds, our method improves both accuracy and efficiency. See Appendix~\ref{app:comp_efficiency} for comparison on FutureX-Past. These comparisons suggest that assigning specific, planner-chosen roles utilizes tool calls much more effectively than repeatedly running the same baseline agent.

\subsection{Live Results with FutureX Leaderboard}
\label{sec:exp-live}

We submitted our framework to four consecutive FutureX weekly rounds from 2026~Apr.~W3 to 2026~May~W2. It is listed on the public leaderboard\footnote{Results are taken from the public leaderboard at \url{https://futurex.live/}, accessed on 2026-05-24.} as \textbf{FutureGPTv1} and referred to as Ours below. We compare against MiroMind's MiroFlow~\cite{miroflow} and the H2O AI Super Agent v1.82\footnote{System blog from H2O.ai: \url{https://h2o.ai/blog/2026/h2oai-super-agent-tops-futurex-leaderboard/}.}. We choose these two systems because they are strong public leaderboard submissions with accompanying public documentation. Full weekly scores, event counts, ranks, and per-level results are reported in Appendix~\ref{app:live-fourweeks}.

\textbf{Live benchmark results have visible week-to-week variation.} FutureX refreshes its question pool every week, so each round has a different set of events, difficulty mix, and resolution status. For this reason, raw scores are not directly comparable across different weeks. Figure~\ref{fig:live-rank-trajectory} therefore presents the public rank results rather than only the absolute score. This view better reflects the live leaderboard setting, where the main comparison is within the same weekly round.

\textbf{Our system has the best average rank.} Across the four submitted rounds, Ours ranks \#4, \#2, \#1, and \#6, giving an average rank of $3.25$. H2O averages $4.50$ over the same four rounds, and MiroFlow averages $5.67$ over the first three rounds. Although forecasting events vary by week and performance fluctuates, our system stays in the leading group throughout the period and achieves the strongest average rank among these public baselines. FutureX also reports an official eight-week overall leaderboard based on a sliding-window Bradley--Terry (BT) model and restricted to agent frameworks with at least three weekly participations. As of June 17, 2026, ours ranks first on this overall leaderboard with a BT rating of $4.61$ (see Appendix Figure~\ref{fig:live-overall}). 

\textbf{Offline scores are consistent with live performance.} Figure~\ref{fig:live-offline-consistency} compares each local run's offline FutureX-Past score with TM with the corresponding method's live score in 2026~May~W2. The 11 points consist of three runs each for Ours, GPT-5.4-medium, and Kimi-K2.5, plus one run each for SC@3 and SC@5. The relationship is strongly positive, with Pearson $r=0.92$ and $R^2=0.85$. Consequently, the performance achieved with our Agentic Time Machine infrastructure closely mirrors actual forecasting capability under genuine real-world uncertainty, establishing a reliable sandbox for LLM prediction evaluation.

\section{Conclusion}
\label{sec:conclusion}

Future-event forecasting is a strong test for LLM agents, but it is hard to evaluate efficiently. Live benchmarks are realistic but slow, while historical replay is fast but vulnerable to answer leakage from the modern web. We propose Agentic Time Machine, an LLM-based infrastructure that wraps web tools with a leakage filter and blocks post-cutoff or answer-revealing content before it reaches the agent. This enables faster and less contaminated local replay while keeping open-ended web interaction. We also propose a planner-solver-aggregator framework that splits each question into several analytical angles, gathers evidence in parallel, and reconciles the results into a final forecast. Across FutureX-Past, Polymarket, and the live FutureX leaderboard, our results show that the system is competitive with strong contemporary forecasting agents and improves over strong single agents.

\bibliography{references}

\appendix

\section{Time Machine: Mechanics and Empirical Behavior}
\label{app:tm}

This section extends Section~\ref{sec:bench} with the implementation and more empirical analysis of Time Machine on the 61-question controlled subset. Section~\ref{app:tm-mechanics} describes how the same filtering approach is applied to our two web tools and presents the core rules for each prompt. The quantitative results in \S\ref{app:tm-perlevel} and \S\ref{app:tm-overfilter}, together with the case studies in \S\ref{app:tm-cases}, are computed on the same paired runs (with and without Time Machine, 5 backbones $\times$ 3 runs) that underlie Table~\ref{tab:tm}.

\subsection{Web Tools Utilized in our System}\label{app:web_tools}
Our forecasting agents utilize two retrieval tools. The \texttt{search} tool executes a Google query and returns a list of result snippets. Each snippet includes a title, a URL, a short summary text, and a publish date if available. The \texttt{visit} tool fetches the body text of a specified URL, which can also summarize the content based on a goal query provided by the agent. Combined, these two tools cover both paths where leaked information might reach the agent. These paths are the brief snippets in the search results and the full page content fetched during a follow-up call.

Both tools are integrated into the agent system using standard JSON function calling schemas. 
\begin{tcolorbox}[
  enhanced, breakable,
  colback=yellow!2!white, colframe=black!40!white,
  colbacktitle=black!15!white, coltitle=black,
  title={\texttt{search} tool schema},
  fonttitle=\small\bfseries,
  arc=2pt, boxrule=0.6pt,
  left=6pt, right=6pt, top=4pt, bottom=4pt,
  toptitle=2pt, bottomtitle=2pt,
]
\begin{Verbatim}[fontsize=\footnotesize]
{
  "name": "search",
  "description": "Batched web search.
    Supply an array 'query'; the tool
    retrieves the top 10 results for
    each query in one call.",
  "parameters": {
    "type": "object",
    "properties": {
      "query": {
        "type": "array",
        "items": {"type": "string"},
        "description": "Array of query
          strings. Include multiple
          complementary search queries
          in a single call."
      }
    },
    "required": ["query"]
  }
}
\end{Verbatim}
\end{tcolorbox}

\begin{tcolorbox}[
  enhanced, breakable,
  colback=yellow!2!white, colframe=black!40!white,
  colbacktitle=black!15!white, coltitle=black,
  title={\texttt{visit} tool schema},
  fonttitle=\small\bfseries,
  arc=2pt, boxrule=0.6pt,
  left=6pt, right=6pt, top=4pt, bottom=4pt,
  toptitle=2pt, bottomtitle=2pt,
]
\begin{Verbatim}[fontsize=\footnotesize]
{
  "name": "visit",
  "description": "Visit webpage(s) and
    return the summary of the content.",
  "parameters": {
    "type": "object",
    "properties": {
      "url": {
        "type": ["string", "array"],
        "items": {"type": "string"},
        "minItems": 1,
        "description": "The URL(s) of the
          webpage(s) to visit. Can be a
          single URL or an array of URLs."
      },
      "goal": {
        "type": "string",
        "description": "The goal of the
          visit, used to focus the
          per-URL summary."
      }
    },
    "required": ["url", "goal"]
  }
}
\end{Verbatim}
\end{tcolorbox}

\subsection{Filter Mechanism and Prompts}
\label{app:tm-mechanics}
Time Machine is a tool-output filtering idea, not a fixed pipeline, and serves as the evaluation infrastructure. The leakage filter is agnostic to the specific tool design and can wrap the output of any retrieval or browsing tool before the agent receives the response. In our implementation, the agent has two web tools, \texttt{search} and \texttt{visit}, so we apply the same LLM-based leakage filter to both tool outputs. The filter follows the two principles:

\begin{itemize}
  \setlength\itemsep{0pt}
  \item \textbf{Direct answer leak}: drop any input whose content directly states or reveals the answer to the prediction question, regardless of date.
  \item \textbf{Post-date info leak}: drop any input that carries a time signal (publish date, URL date, in-text event date) later than the event's cutoff time, regardless of content.
\end{itemize}

The filter applies both principles wherever a tool returns web content. For the \texttt{search} tool, it checks the returned titles and snippets. For the \texttt{visit} tool, it inspects the fetched page body. This comprehensive filtering guards against two common vulnerabilities. First, an innocuous search title might hide an answer-bearing webpage body. Second, a missing snippet date could be later revealed by an in-page timestamp. Because a leak that is invisible in a brief snippet may still appear in the full page content if the agent visits that URL, checking both tool outputs substantially reduces leakage.

Table~\ref{tab:tm-tool-outputs} contrasts the two tool-output filters along the dimensions where they differ: the LLM and what the agent sees when the filter blocks. We then list the core rules of each prompt.

\begin{table}[h]
  \centering
  \small
  \setlength{\tabcolsep}{4pt}
  \begin{tabular}{p{2.5cm}p{6.3cm}p{6.5cm}}
    \toprule
    & \textbf{Search output} & \textbf{Visit output} \\
    \midrule
    Filter LLM & Kimi-K2.5 & GPT-5.2 \\
    Input per call & query, time range, problem, ground truth, $\leq 10$ snippets & problem, ground truth, cutoff, fetched page body \\
    Effect of block & snippet omitted from the tool's returned list & tool returns \texttt{[ACCESS RESTRICTED]} \\
    \bottomrule
  \end{tabular}
  \caption{How Time Machine is applied to the two web-tool outputs in our system. The search-output filter processes short snippets in batches; the visit-output filter processes a full page body (up to roughly 12K characters) and is therefore more expensive per call.}
  \label{tab:tm-tool-outputs}
\end{table}

We use different filter LLMs for the two tool outputs as a speed and cost trade-off. The agent calls \texttt{search} much more often than \texttt{visit}, and each search call may contain up to ten snippets. We therefore use Kimi-K2.5 for search outputs, since it is faster and cheaper while still strong enough for short snippets. The \texttt{visit} tool is called less, but each call contains a much longer page body and can require more careful reading. We therefore use advanced GPT-5.2 for visit outputs. This model choice is an implementation decision; the Time Machine idea only requires that each tool output be checked against the same two leakage principles.

\definecolor{tmpromptbg}{HTML}{FAF4FB}
\definecolor{tmprompttitle}{HTML}{C2BBF0}
\definecolor{tmpromptframe}{HTML}{62BFED}

\paragraph{Search-side filter.} The search-side prompt supplies the judge with the Google \texttt{tbs} time-range string, the current real-world date, the prediction problem, the ground truth, and the numbered list of result snippets (title, link, snippet text, and a published date when one is available). Its key rules are:

\begin{tcolorbox}[
  enhanced, breakable,
  colback=tmpromptbg, colframe=tmpromptframe,
  colbacktitle=tmprompttitle, coltitle=black,
  title={Search-side filter rules},
  fonttitle=\small\bfseries,
  arc=2pt, boxrule=0.6pt,
  left=6pt, right=6pt, top=4pt, bottom=4pt,
  toptitle=2pt, bottomtitle=2pt,
]
\begin{Verbatim}[fontsize=\footnotesize, breaklines, breakanywhere, breaksymbolleft={}, breaksymbolright={}]
Drop a result if ANY rule applies:

1) Time: The result has a clear published date, URL date, title date, or snippet date outside the allowed cd_min-cd_max window.
2) Spoiler: The title or snippet directly resolves the prediction task. This applies even when the result has no reliable date, and even when the answer appears only in the snippet rather than the full page.
3) Hindsight: The snippet uses result/reporting language that makes a still-protected future outcome appear settled, known, completed, reported, won, lost, announced, confirmed, or published.

Exception: keep background, previews, speculation, or historical context inside the allowed time window that does not resolve the task.
\end{Verbatim}
\end{tcolorbox}

\paragraph{Visit-side filter.} The visit-side prompt supplies the judge with the prediction problem, the ground truth, the cutoff time, the page source URL, and the page body (truncated to roughly 12K characters). Its key rules are:

\begin{tcolorbox}[
  enhanced, breakable,
  colback=tmpromptbg, colframe=tmpromptframe,
  colbacktitle=tmprompttitle, coltitle=black,
  title={Visit-side filter rules},
  fonttitle=\small\bfseries,
  arc=2pt, boxrule=0.6pt,
  left=6pt, right=6pt, top=4pt, bottom=4pt,
  toptitle=2pt, bottomtitle=2pt,
]
\begin{Verbatim}[fontsize=\footnotesize, breaklines, breakanywhere, breaksymbolleft={}, breaksymbolright={}]
Rules:
1. Return "LEAKED" if the content directly states, strongly implies, or confirms the answer to the prediction question.
2. Return "OUT_OF_TIME" only if the content itself contains a clear time signal (publish/update/event/result time) that is later than the cutoff time_point.
3. Return "SAFE" if the content is only background, historical context, previews, speculation, or the time signal is missing/uncertain.
\end{Verbatim}
\end{tcolorbox}

\subsection{Per-level effect of Time Machine}
\label{app:tm-perlevel}

The aggregate Time Machine drop in Table~\ref{tab:tm} falls mostly on the retrieval-heavy levels. Every backbone loses $30$ to $50$ points on Level 4 (quantitative) but only $10$ to $28$ points on Level 1 (few options). Figure~\ref{fig:tm-per-level} (Section~\ref{sec:exp-timemachine}) plots these per-level $\Delta$ values; Table~\ref{tab:tm-perlevel-abs} reports the three-run means and standard deviations underlying the figure.

\begin{table*}[h]
  \centering
  \small
  \setlength{\tabcolsep}{2.2pt}
  \begin{tabular}{l cc cc cc cc cc}
    \toprule
    & \multicolumn{2}{c}{L1} & \multicolumn{2}{c}{L2} & \multicolumn{2}{c}{L3} & \multicolumn{2}{c}{L4} & \multicolumn{2}{c}{Weighted} \\
    \cmidrule(lr){2-3}\cmidrule(lr){4-5}\cmidrule(lr){6-7}\cmidrule(lr){8-9}\cmidrule(lr){10-11}
    Model & w/o & w/ & w/o & w/ & w/o & w/ & w/o & w/ & w/o & w/ \\
    \midrule
    Kimi-K2.5         & 90.2$_{\pm 5.6}$ & 68.6$_{\pm 2.8}$ & 87.4$_{\pm 2.3}$ & 53.7$_{\pm 4.0}$ & 30.8$_{\pm 1.3}$ &  7.9$_{\pm 1.4}$ & 62.2$_{\pm 1.5}$ & 17.9$_{\pm 2.9}$ & 60.6$_{\pm 0.3}$ & 27.1$_{\pm 1.7}$ \\
    GLM-5             & 96.1$_{\pm 2.8}$ & 68.2$_{\pm 7.2}$ & 89.4$_{\pm 2.9}$ & 58.2$_{\pm 6.0}$ & 42.2$_{\pm 2.1}$ &  8.4$_{\pm 4.3}$ & 68.3$_{\pm 2.3}$ & 24.9$_{\pm 2.0}$ & 67.5$_{\pm 1.4}$ & 31.0$_{\pm 2.3}$ \\
    DeepSeek-V4-Flash & 86.3$_{\pm 2.8}$ & 68.6$_{\pm 7.3}$ & 85.9$_{\pm 0.5}$ & 64.0$_{\pm 1.1}$ & 48.9$_{\pm 3.4}$ & 12.4$_{\pm 5.0}$ & 71.6$_{\pm 1.4}$ & 29.1$_{\pm 4.4}$ & 69.1$_{\pm 0.9}$ & 35.0$_{\pm 1.4}$ \\
    GPT-5.4-medium    & 80.4$_{\pm 2.8}$ & 70.0$_{\pm 0.9}$ & 85.9$_{\pm 0.2}$ & 66.8$_{\pm 4.4}$ & 34.6$_{\pm 3.4}$ & 12.6$_{\pm 5.2}$ & 72.0$_{\pm 4.5}$ & 22.5$_{\pm 1.2}$ & 64.4$_{\pm 2.6}$ & 33.1$_{\pm 1.1}$ \\
    Claude-4.6-Sonnet & 88.2$_{\pm 4.8}$ & 71.9$_{\pm 3.3}$ & 83.5$_{\pm 3.1}$ & 43.2$_{\pm 6.0}$ & 18.0$_{\pm 1.4}$ &  6.9$_{\pm 2.8}$ & 42.2$_{\pm 0.7}$ & 12.2$_{\pm 1.0}$ & 47.8$_{\pm 0.6}$ & 22.8$_{\pm 1.0}$ \\
    \bottomrule
  \end{tabular}
  \caption{Per-level score with Time Machine off (w/o) and on (w/), reported as mean $\pm$ standard deviation across 3 runs. The $\Delta$ bars and error bars in Figure~\ref{fig:tm-per-level} are computed from these values.}
  \label{tab:tm-perlevel-abs}
\end{table*}

\definecolor{tmcasewo}{HTML}{8FB8ED}  
\definecolor{tmcasew}{HTML}{C2BBF0}   

\subsection{Case studies}
\label{app:tm-cases}

We present two representative cases from the paired GPT-5.4-medium runs that underlie Tables~\ref{tab:tm} and~\ref{tab:tm-perlevel-abs}.

\paragraph{Case A: AAPL daily high (L4 quantitative, qid \texttt{6851736beb11c800614780df}).}
\textbf{Question:} "2026-01-23, what will the high of Apple stock (AAPL) be for the day (in US\$)?" \textbf{GT:} 249.41. The w/o TM agent reproduces $\boxed{249.41}$ exactly under all three trials; the w/ TM agent instead answers 250.90, 250.92, and 252.40. Table~\ref{tab:tm-case-aapl} traces the second trail of w/ TM and w/o TM. Both filter stages are visible: the search-side filter silently shrinks the result list, and the visit-side filter blocks two of three data sources the agent tries to read.

\begin{table}[!ht]
  \centering
  \small
  \setlength{\tabcolsep}{5pt}
  \renewcommand{\arraystretch}{1.15}
  \begin{tabular}{cp{15cm}}
    \toprule
    \# & \textbf{Step} \\
    \midrule
    \rowcolor{tmcasewo}
    \multicolumn{2}{l}{\textbf{w/o Time Machine}} \\
    1 & \texttt{search}: "AAPL historical Jan 23 2026 high" (+ 2 parallel reformulations) \\
    2 & $\hookrightarrow$ 10 results; the Yahoo Finance result includes the snippet "Jan 23, 2026, 247.32, 249.41, 244.68, 248.04" (OHLC row, high in plain sight) \\
    3 & \textbf{Answer: $\boxed{249.41}$} \quad \emph{(exact match)} \\
    \midrule
    \rowcolor{tmcasew}
    \multicolumn{2}{l}{\textbf{w/ Time Machine}} \\
    1 & \texttt{search}: "AAPL Jan 2026 news / price / analyst" (+ 2 parallel reformulations) \\
    2 & $\hookrightarrow$ 10 results, all pre-cutoff news / analyst pieces; no Jan 23 OHLC anywhere in the snippets \\
    3 & \texttt{search}: "AAPL Jan 22 2026 close high low" (+ 2 parallel reformulations) \\
    4 & $\hookrightarrow$ 6 results returned (the search-side filter silently dropped 4 of 10; no Jan 23 OHLC in the surviving snippets) \\
    5 & \texttt{visit} ($\times 3$ in parallel): \\
    & \texttt{investor.apple.com/stock-price},\\
    & \texttt{stockanalysis.com/stocks/aapl/history/},\\
    & \texttt{nasdaq.com/\ldots/aapl/historical} \\
    6 & $\hookrightarrow$ \texttt{investor.apple.com}: \texttt{[ACCESS RESTRICTED]} \\
    7 & $\hookrightarrow$ \texttt{stockanalysis.com}: \texttt{[ACCESS RESTRICTED]} \\
    8 & $\hookrightarrow$ \texttt{nasdaq.com}: page passes the filter; its content has Jan 22 OHLC (high 251.00) but no Jan 23 row \\
    9 & \textbf{Answer: $\boxed{250.92}$} \quad \emph{(wrong; extrapolated from Jan 22 high 251.00, off by 1.51 from GT)} \\
    \bottomrule
  \end{tabular}
  \caption{Step-by-step AAPL daily high transcript (Level~4, GPT-5.4-medium run 2); step numbers restart per regime. Runs 1 and 3 get the final answer of 250.90 and 252.40, respectively.}
  \label{tab:tm-case-aapl}
\end{table}

\paragraph{Case B: Nikkei 225 daily close (L4 quantitative, qid \texttt{6851741059f71f006037a1ee}).}
\textbf{Question:} "2026-01-23, what will be the Nikkei Stock Average (Nikkei 225)'s close for the day?" \textbf{GT:} 53,846.87 (agent cutoff: 2026-01-22). The w/o TM agent reproduces $\boxed{53{,}846.87}$ exactly under all three runs; the w/ TM agent instead answers 54,110.50, 53,796.27, and 54,172.09. Table~\ref{tab:tm-case-nikkei} traces the run-1 transcript. Without Time Machine, the first search returns the Yahoo Finance historical-prices page as the top result. Its snippet contains the Jan 23 OHLC row with the close (53,846.87) in plain sight, and the agent answers from the snippet without calling \texttt{visit}. With Time Machine, the answer-bearing sources are all blocked. A third-party LinkedIn post quoting a different day's Nikkei close (54,110.5, undated in the snippet) nevertheless slips through, and the agent anchors on this misattributed value.

\begin{table}[!ht]
  \centering
  \small
  \setlength{\tabcolsep}{5pt}
  \renewcommand{\arraystretch}{1.15}
  \begin{tabular}{cp{15cm}}
    \toprule
    \# & \textbf{Step} \\
    \midrule
    \rowcolor{tmcasewo}
    \multicolumn{2}{l}{\textbf{w/o Time Machine}} \\
    1 & \texttt{search}: "Nikkei 225 historical Jan 23 2026 close" (+ 2 parallel reformulations) \\
    2 & $\hookrightarrow$ 10 results; the Yahoo Finance result includes the snippet "Jan 23, 2026, 53,898.45, 54,050.84, 53,603.68, 53,846.87" (OHLC row, close in plain sight) \\
    3 & \textbf{Answer: $\boxed{53{,}846.87}$} \quad \emph{(exact match)} \\
    \midrule
    \rowcolor{tmcasew}
    \multicolumn{2}{l}{\textbf{w/ Time Machine}} \\
    1 & \texttt{search}: "Nikkei 225 close Jan 22 2026 Reuters" (+ 4 parallel reformulations) \\
    2 & $\hookrightarrow$ 8 results, all pre-cutoff macro articles; no Jan 23 Nikkei close anywhere in the snippets \\
    3 & \texttt{search}: 4 more Jan 23 Nikkei Reuters queries \\
    4 & $\hookrightarrow$ 9 results returned, among them a LinkedIn post snippet quoting "The Nikkei 225 declined 0.42\% to 54,110.5" with no per-day date \\
    5 & \texttt{visit}: \\
    & \texttt{reuters.com/world/china/global-\ldots},\\
    & \texttt{cnbc.com/2026/01/23/asia-pacific-\ldots} \\
    6 & $\hookrightarrow$ \texttt{reuters.com}: \texttt{[ACCESS RESTRICTED]} \\
    7 & $\hookrightarrow$ \texttt{cnbc.com}: \texttt{[ACCESS RESTRICTED]} \\
    8 & \texttt{search}: additional targeted queries using \texttt{site:reuters.com} \\
    9 & $\hookrightarrow$ 4 results, all unrelated old articles; no Jan 23 Nikkei data \\
    10 & \texttt{search}: exact-match "\texttt{54,110.5}" Nikkei 225 Jan 23 2026 confirmation \\
    11 & $\hookrightarrow$ 5 results returned, re‑quoting the same undated LinkedIn snippet \\
    12 & \textbf{Answer: $\boxed{54{,}110.50}$} \quad \emph{(wrong; anchored on undated LinkedIn snippet, off by 263.63 from GT)} \\
    \bottomrule
  \end{tabular}
  \caption{Step-by-step Nikkei 225 daily-close transcript (Level~4, GPT-5.4-medium run 1); step numbers restart per regime. Runs 2 and 3 get the final answer of 53,796.27 and 54,172.09, respectively.}
  \label{tab:tm-case-nikkei}
\end{table}

\subsection{Filter validation and the over-block bias}
\label{app:tm-overfilter}

LLM-based leakage filters face a trade-off between over-blocking and under-blocking. Over-blocking removes legitimate evidence and depresses scores for the wrong reasons; under-blocking lets post-cutoff content reach the agent. To measure where our filter sits on this trade-off, we sampled 150 search-output calls and 150 visit-output calls from one full run (61 questions $\times$ 4 levels, GPT-5.4-medium, TM on). Each sample was labeled as $\mathit{Y}$ (should block: snippet or page directly resolves the question, or carries post-cutoff content) or $\mathit{N}$ (should keep: pre-cutoff background, preview, forecast, or unrelated context). Labels were assigned blind to the filter's decision. Table~\ref{tab:tm-filter-validation} reports counts and metrics for the two tool outputs, with $\mathit{Y}$ as the positive class: TP = blocked among $\mathit{Y}$; FN = kept among $\mathit{Y}$ (a missed leak); FP = blocked among $\mathit{N}$ (an over-block); TN = kept among $\mathit{N}$. P, R, F1 denote precision, recall, and their harmonic mean.

\begin{table}[h]
  \centering
  \setlength{\tabcolsep}{3pt}
  \begin{tabular}{l *{7}{c}}  
    \toprule
    Tool & TP & FN & FP & TN & P & R & F1 \\
    \midrule
    Search  & 24 & 2 & 4  & 120 & 0.857 & 0.923 & 0.889 \\
    Visit   & 41 & 0 & 14 & 95  & 0.745 & \textbf{1.000} & 0.854 \\
    \bottomrule
  \end{tabular}
  \caption{Filter validation on 150 labeled samples per tool output. The filter is asymmetric. At the page level, it never misses a true leak in our sample (recall 1.000), at the cost of a precision of 0.745, meaning roughly one in four blocked pages was actually safe. At the snippet level, it misses 2 of 26 leaks (recall 0.923).}
  \label{tab:tm-filter-validation}
\end{table}

The filter is biased toward over-blocking by design: block-when-in-doubt. The remainder of this subsection makes the trade-off concrete via one case from each side, then explains why we prefer this direction.

\paragraph{A miss (false negative).} The question "2026-03-19, which movie will be the latest \#1 in Box Office Mojo's Domestic Box Office Weekly?" has ground truth \emph{Hoppers}. For one of the agent's queries, Google returned a result with \texttt{published\_date: Apr 17, 2019} (a stale or spoofed CMS field). Its snippet read: "\texttt{Box Office -- March 13--15, 2026: HOPPERS, REMINDERS OF HIM, UNDERTONE, \& More March 16, 2026.}" The filter prompt covers this case in two rules: Time ("a date in the page path later than \texttt{cd\_max} $\Rightarrow$ drop") and Spoiler ("the snippet directly resolves the question $\Rightarrow$ drop"). The LLM judge missed both. It failed to identify the out-of-range date embedded in the page path. It also missed the spoiler, because reading "March 13--15 \#1 = HOPPERS" as resolving the March 19 weekly chart requires knowing that BOM's weekly chart reflects the prior weekend. \emph{Consequence:} a single line of search output names the answer. The agent has no incentive to issue further queries, fetch any page, or reason at all. The prediction task collapses to copying the first capitalized title in the snippet.

\paragraph{An over-block (false positive).} The question "Will Bugonia get an Oscar nomination for Best Picture?" has ground truth Yes (\texttt{cd\_max} = 2026-01-22; the Academy announced nominations on January 23). Two days before the announcement, Variety published \emph{Final Academy Awards Nomination Predictions 2026}. The article listed each category's projected nominees, with Bugonia in the projected Best Picture lineup. The visit-side filter returned \texttt{LEAKED} with the reason "Lists Bugonia in Best Picture final nomination predictions, strongly implying it will be nominated." This is a hard call. The Variety prediction does approach answering the question. But it is also exactly the kind of pre-cutoff editorial signal a forecasting agent should weigh. The filter chose the conservative interpretation. \emph{Consequence:} the agent loses one prediction source. It then issues additional queries that surface parallel signals (Awards Watch and Hollywood Reporter's own forecasts, already-announced Golden Globe and BAFTA longlist outcomes, Critics Choice nominations) and reassembles equivalent context for Bugonia. The total signal the agent can weigh converges to roughly the same direction; only the search-and-visit budget is higher.

\paragraph{Why over-blocking is the safer direction.} The two failure modes have asymmetrical effects on the agent. A false negative \emph{terminates} reasoning: the leaked snippet (as in the Hoppers example) is itself the answer, and the agent rationally stops searching once it sees one. A false positive only \emph{lengthens} reasoning. The dropped source is usually one of many redundant copies of the same pre-cutoff context that the web carries about most questions, and the agent recovers the missing signal by issuing more queries or visiting parallel pages. In the limit, a false positive costs search budget; a false negative costs the entire scientific value of the w/ TM run, because the agent never had to predict. We therefore tune the filter toward the first failure mode. The precision cost in Table~\ref{tab:tm-filter-validation} is the price of treating "block-when-in-doubt" as a safe default for a contamination filter.

\section{Local Datasets}
\label{app:data}
This appendix introduces the two resolved local datasets used in our offline experiments. The FutureX-Past subset is used for the Time Machine validation and controlled local comparison in the main paper. The Polymarket subset adds a separate high-difficulty multiple-choice forecasting set. Both subsets are formatted as pre-resolution prediction tasks; ground truth is used only for offline scoring.

\definecolor{databoxbg}{HTML}{FFFFFF}
\definecolor{databoxtitle}{HTML}{E6E6E6}
\definecolor{databoxframe}{HTML}{B8B8B8}

\begin{figure}[t]
  \centering
  \includegraphics[width=0.9\columnwidth]{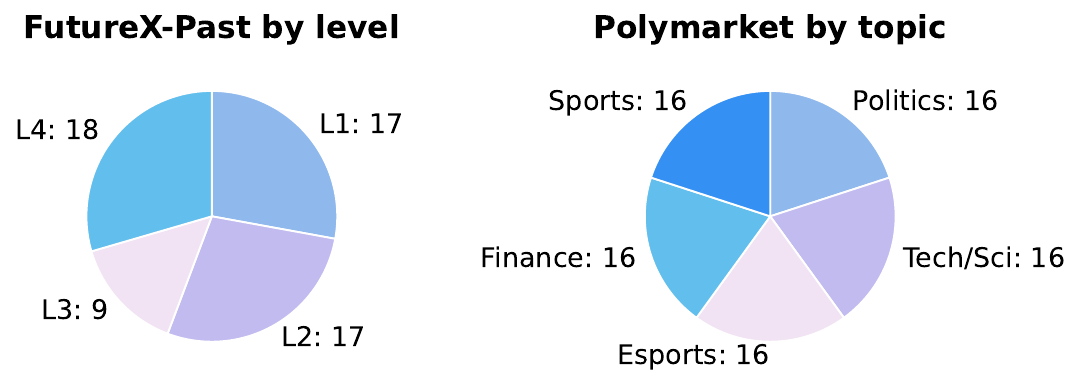}
  \caption{Observed composition of the two local subsets. FutureX-Past covers all four benchmark levels, while the Polymarket subset contains exactly 16 items from each of five topics.}
  \label{fig:local-data-distribution}
\end{figure}

\begin{table}[t]
  \centering
  \small
  \setlength{\tabcolsep}{4pt}
  \renewcommand{\arraystretch}{1.12}
  \begin{tabular}{@{}p{3cm}p{12.4cm}@{}}
    \toprule
    \textbf{Subset} & \textbf{Construction summary} \\
    \midrule
    FutureX-Past & 61 resolved FutureX historical questions, sampled from \url{https://huggingface.co/datasets/futurex-ai/Futurex-Past}, observed level counts are 17/17/9/18 for L1-L4. \\
    Polymarket & 80 resolved Polymarket questions sampled from a 4,573-item balanced parent pool built from the public Gamma API; the subset is evenly split across five topics. \\
    \bottomrule
  \end{tabular}
  \caption{High-level summary of the two local evaluation subsets.}
  \label{tab:local-data-summary}
\end{table}

\subsection{FutureX-Past}
\label{app:data-futurex}

FutureX-Past is the historical archive of the FutureX benchmark. Its entries are resolved prediction tasks with a question prompt, ground truth, resolution time, and difficulty level. Because these questions are already resolved, they are not used as evidence of live forecasting ability by themselves.

The FutureX-Past subset contains 61 questions. The observed level distribution is 17/17/9/18 for Levels 1--4 respectively (Figure~\ref{fig:local-data-distribution}). Level 1 mostly consists of binary or small categorical questions, Level 2 contains larger multiple-choice questions, Level 3 contains open-ended ranking or compositional questions, and Level 4 contains quantitative or open-ended questions scored by the official graded metric. The subset spans resolution dates from 2026-01-15 to 2026-04-06.

\begin{tcolorbox}[
  enhanced, breakable,
  colback=databoxbg, colframe=databoxframe,
  colbacktitle=databoxtitle, coltitle=black,
  title={Representative FutureX-Past examples},
  fonttitle=\small\bfseries,
  arc=2pt, boxrule=0.6pt,
  left=6pt, right=6pt, top=4pt, bottom=4pt,
  toptitle=2pt, bottomtitle=2pt,
]
\footnotesize
\textbf{L1.} Islanders vs. Oilers. \emph{GT:} \texttt{Islanders}.\\
\textbf{L2.} Q1 S\&P 500 Performance. \emph{GT:} \texttt{A}.\\
\textbf{L3.} 2026-03-19, value in Chinese Yuan of 100 units of British Pound at the central parity rate. \emph{GT:} \texttt{915.04}.\\
\textbf{L4.} 2026-01-23, high of Apple stock (AAPL) for the day, in US\$. \emph{GT:} \texttt{249.41}.
\end{tcolorbox}

\paragraph{Local corrections.}
We compared the 61 local rows against the corresponding rows in \url{https://huggingface.co/datasets/futurex-ai/Futurex-Past}. Most differences are formatting-only changes in prompt whitespace. Three items were manually corrected in the local subset. Two are corrected ground-truth labels for multiple-choice questions; the third fixes a mismatch between the required answer format in the prompt and the stored ground-truth label.

\begin{tcolorbox}[
  enhanced, breakable,
  colback=databoxbg, colframe=databoxframe,
  colbacktitle=databoxtitle, coltitle=black,
  title={Manual corrections in the FutureX-Past subset},
  fonttitle=\small\bfseries,
  arc=2pt, boxrule=0.6pt,
  left=6pt, right=6pt, top=4pt, bottom=4pt,
  toptitle=2pt, bottomtitle=2pt,
]
\footnotesize
\textbf{\texttt{6957ba8a03568a006853e828}.} For "How will the winner of the Portugal presidential election win?", the ground truth is corrected from \texttt{A} (only candidate above 50\% in round one) to \texttt{B} (wins first round and the runoff).\\[2pt]
\textbf{\texttt{69b7f816d596fb005d43a317}.} For "F1: Who will win the 2026 Aramco Japanese Grand Prix?", the ground truth is corrected from \texttt{A} (George Russell) to \texttt{B} (Kimi Antonelli).\\[2pt]
\textbf{\texttt{69624695e87498005daa01d6}.} The prompt asks for \texttt{\textbackslash boxed\{Succeeded\}} or \texttt{\textbackslash boxed\{Backfired\}}, but the stored ground truth was \texttt{A}; we correct it to \texttt{Succeeded} to match the prompt's literal answer format.
\end{tcolorbox}

\subsection{Polymarket}
\label{app:data-polymarket}

The Polymarket dataset is constructed from resolved markets and events collected through Polymarket's public Gamma API. We used the keyset-paginated event endpoint because Polymarket events often group several related binary submarkets, which can be naturally converted into multi-option, sometimes multi-answer, prediction questions. The parent collection is restricted to resolved items between 2026-01-01 and 2026-05-08.

For each event, we convert the market or grouped event into a labeled multiple-choice question. Single-market items use the market outcomes as options. Grouped events with multiple Yes/No submarkets use each submarket as one option, preferring concise option text from \texttt{groupItemTitle}, then \texttt{shortOutcomes}, then the market question or slug. Ground truth labels are inferred from final outcome prices: a label is marked correct when the corresponding winning outcome price is at least 0.99. For grouped events, all submarkets whose Yes price crosses this threshold contribute labels to the ground-truth set.

The parent dataset applies filtering and balancing before the 80-item evaluation subset is drawn. The construction keeps only resolved items with at least one confident ground-truth label, filters noisy or high-frequency tags such as recurring and very short crypto markets from the main pool, favors longer-horizon multi-option grouped events, and caps short-horizon and binary supplements so they do not dominate the benchmark. The resulting balanced parent pool has 4,573 items. The 80-item subset is a focused hard set: it has exactly 16 items in each of five topics, no binary items, no short-horizon items, 67 multi-answer items, option counts from 8 to 26, and a mean heuristic difficulty score of 49.7. Table~\ref{tab:polymarket-difficulty-score} summarizes the construction-time heuristic used to rank and balance candidate items.

\begin{table}[t]
  \centering
  \scriptsize
  \setlength{\tabcolsep}{3pt}
  \renewcommand{\arraystretch}{1.08}
  \begin{tabular}{@{}p{2.55cm}p{6.45cm}p{6.35cm}@{}}
    \toprule
    \textbf{Signal} & \textbf{Score rule} & \textbf{Effect} \\
    \midrule
    Option count & $+18$ for $\geq 12$; $+14$ for 8--11; $+10$ for 4--7; $+3$ for $=3$; $-10$ for $\leq 2$ & Rewards larger answer spaces. \\
    Event shape & $+8$ for grouped events; $+6$ for multiple correct labels & Rewards questions that combine several submarkets. \\
    Horizon & $+12$ for $\geq 1$ day; $+4$ for 6--24 hours; $-14$ for $<6$ hours & Downweights very short markets. \\
    Topic/tags & $+8$ for politics, finance/economy, tech/science, or world news; penalties for recurring, up-or-down, and crypto-price tags & Favors less mechanical domains and filters noisy high-frequency markets. \\
    \bottomrule
  \end{tabular}
  \caption{Polymarket difficulty-score heuristic. The score is used only during dataset construction for ranking and balancing candidate items.}
  \label{tab:polymarket-difficulty-score}
\end{table}

\begin{tcolorbox}[
  enhanced, breakable,
  colback=databoxbg, colframe=databoxframe,
  colbacktitle=databoxtitle, coltitle=black,
  title={Representative Polymarket examples},
  fonttitle=\small\bfseries,
  arc=2pt, boxrule=0.6pt,
  left=6pt, right=6pt, top=4pt, bottom=4pt,
  toptitle=2pt, bottomtitle=2pt,
]
\footnotesize
\textbf{Politics.} \textbf{Question:} What will Trump say this week? (April 26)
\textbf{Options:} A. Make America Great Again; B. Transgender; C. Trump Strait / Strait of Trump; D. DoorDash / McDonald's; E. Shit; F. Genius; G. Excursion / Journey; H. The War Is Over; I. Kamala; J. UFC / Dana White; K. Pope / Leo; L. Tough Cookie; M. Death to America; N. My Father / Fred; O. Boy oh boy; P. Crypto / Bitcoin.
\textbf{GT:} \texttt{A, B, F, G, H, I, P}.\\[3pt]
\textbf{Tech/science.} \textbf{Question:} Where will it snow this weekend? (January 24--26)
\textbf{Options:} A. New York City; B. Dallas; C. Miami; D. Seattle; E. Washington D.C.; F. Phoenix; G. Philadelphia; H. Atlanta; I. Boston; J. Pittsburgh.
\textbf{GT:} \texttt{A, B, E, G, I, J}.\\[3pt]
\textbf{Esports/games.} \textbf{Question:} BLAST Open Rotterdam 2026: Which teams make it to the Grand Final?
\textbf{Options:} A. Vitality; B. Furia; C. Mouz; D. Team Falcons; E. Parivision; F. Team Spirit; G. The MongolZ; H. Natus Vincere; I. FaZe; J. Aurora; K. Team Liquid; L. B8; M. NRG; N. Ninjas in Pyjamas; O. Tyloo; P. 9z.
\textbf{GT:} \texttt{A, H}.\\[3pt]
\textbf{Finance/economy.} \textbf{Question:} What will BlackRock say during their next earnings call?
\textbf{Options:} A. Performance Fee; B. Aladdin; C. Digital Asset; D. ETF Inflow; E. Private market; F. Crypto / Bitcoin; G. Record AUM; H. Stablecoin.
\textbf{GT:} \texttt{A, B, C, E}.\\[3pt]
\textbf{Sports.} \textbf{Question:} Angers SCO vs. Paris Saint-Germain FC - Player Props
\textbf{Options:} A. Goalscorer: Ousmane Dembele; B. Goalscorer: Goncalo Ramos; C. Goalscorer: Bradley Barcola; D. Goalscorer: Desire Doue; E. Goalscorer: Khvicha Kvaratskhelia; F. Goalscorer: Lee Kang-in; G. Goalscorer: Senny Mayulu; H. Goalscorer: Ibrahim Mbaye; I. Goalscorer: Fabian Ruiz; J. Goalscorer: Nuno Mendes; K. Goalscorer: Vitinha; L. Goalscorer: Joao Neves; M. Goalscorer: Achraf Hakimi; N. Goalscorer: Amine Sbai; O. Goalscorer: Lanroy Machine; P. Goalscorer: Peter Prosper.
\textbf{GT:} \texttt{F, G}.
\end{tcolorbox}

\section{Additional Experiment Details}
\subsection{Models Configuration}
In our paper, we utilize six large language models: GPT-5.4-medium, Claude-4.6-Sonnet, Gemini-3.1-Pro, Kimi-K2.5, DeepSeek-V4-Flash and GLM-5. The first three of them are closed-source models with no public model size information. For the remaining models, Kimi-K2.5 has 1T total parameters with 32B activated parameters, DeepSeek-V4-Flash has 284B total parameters with 13B activated parameters, and GLM-5 has 744B total parameters with 40B activated parameters.

\subsection{Week-by-week FutureX live leaderboard results}
\label{app:live-fourweeks}

\begin{figure*}[t]
    \centering
    \includegraphics[width=\linewidth]{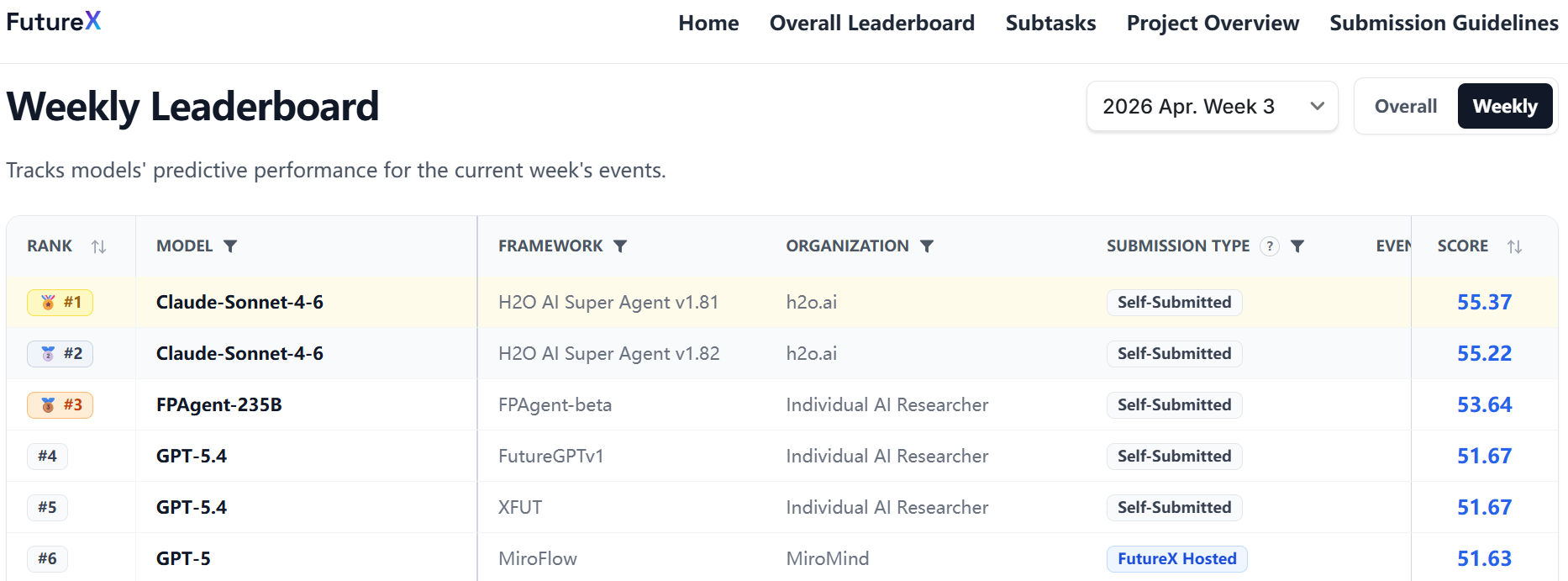}
    \caption{FutureX live leaderboard snapshot for Apr.~W3, 2026, accessed on May 24, 2026.}
    \label{fig:live-aprw3}
\end{figure*}

\begin{figure*}[t]
    \centering
    \includegraphics[width=\linewidth]{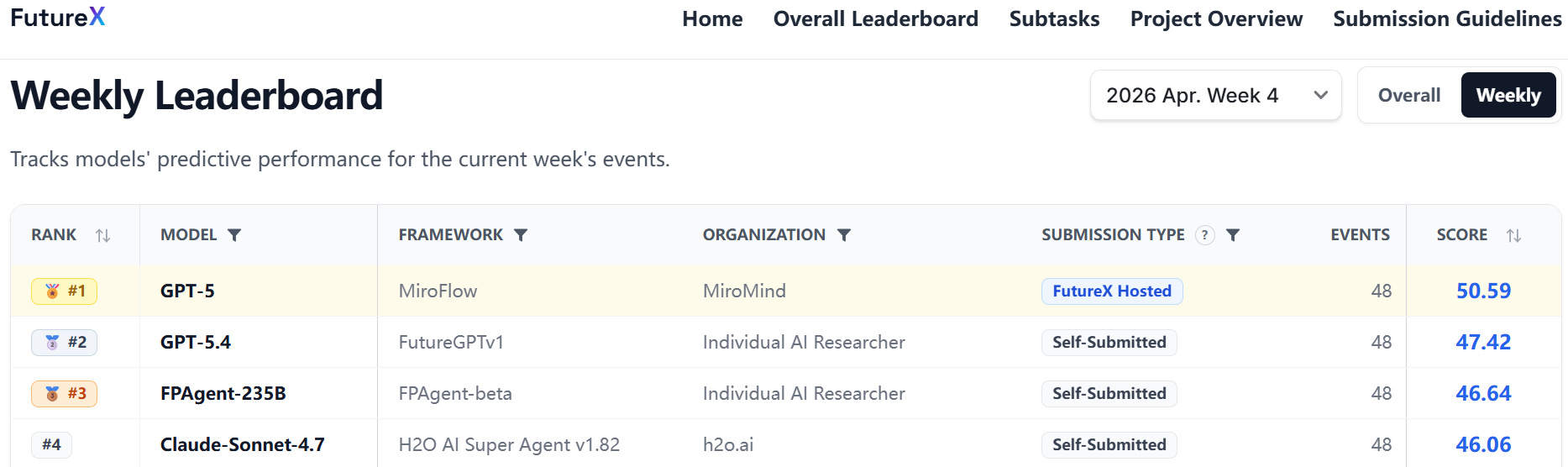}
    \caption{FutureX live leaderboard snapshot for Apr.~W4, 2026, accessed on May 24, 2026.}
    \label{fig:live-aprw4}
\end{figure*}

\begin{figure*}[t]
    \centering
    \includegraphics[width=\linewidth]{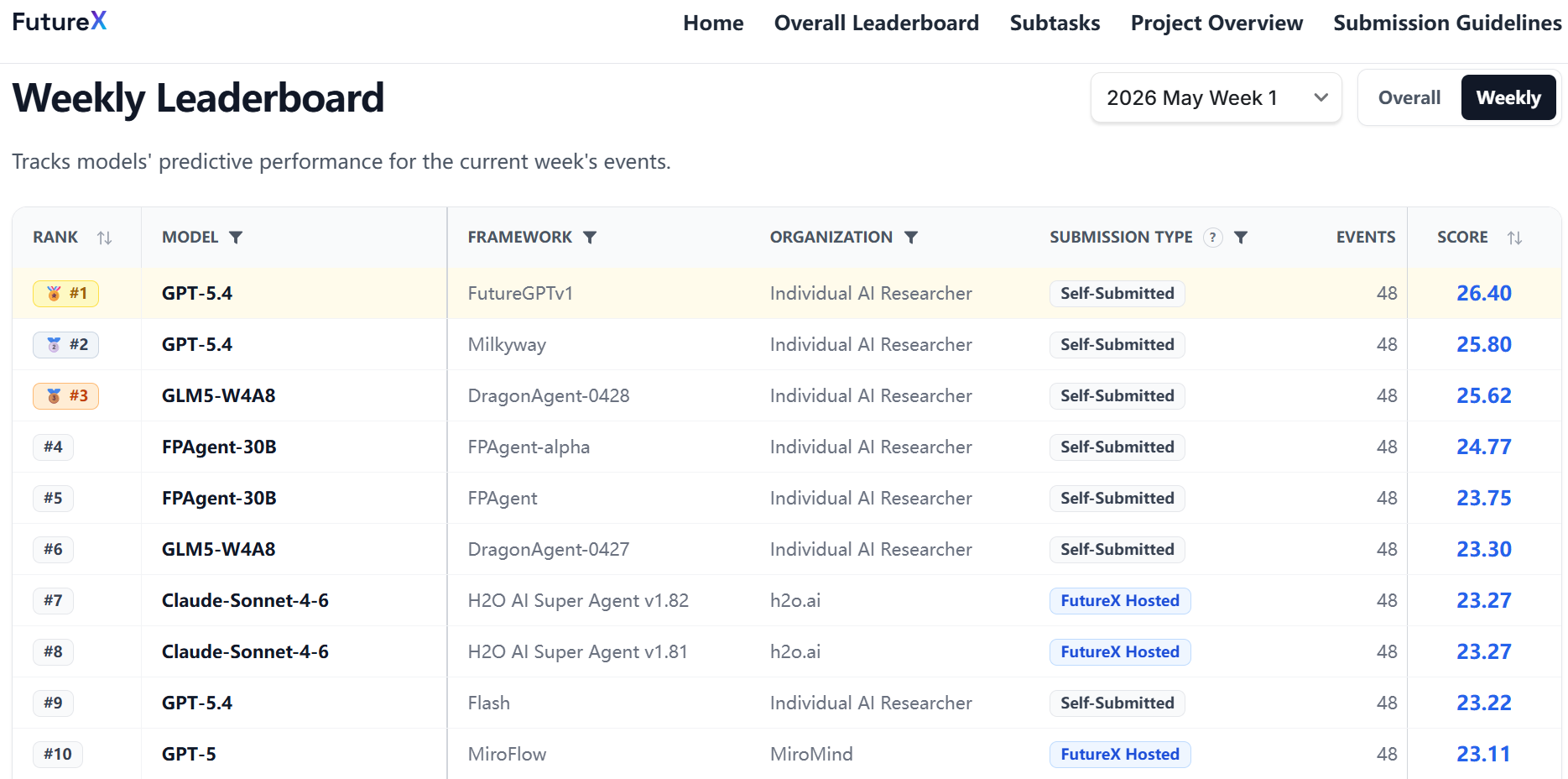}
    \caption{FutureX live leaderboard snapshot for May~W1, 2026, accessed on May 24, 2026.}
    \label{fig:live-mayw1}
\end{figure*}

\begin{figure*}[t]
    \centering
    \includegraphics[width=\linewidth]{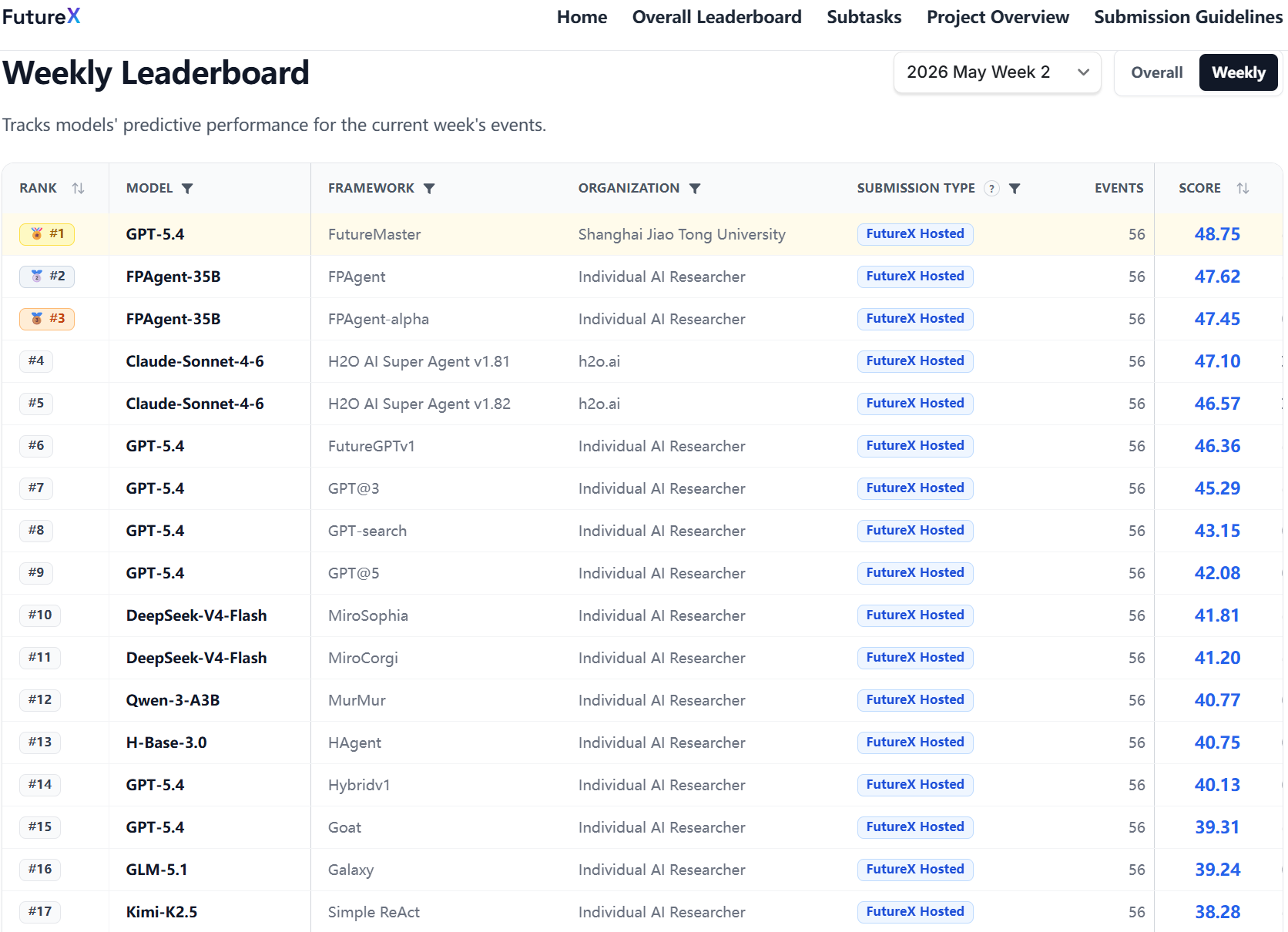}
    \caption{FutureX live leaderboard snapshot for May~W2, 2026, accessed on May 24, 2026.}
    \label{fig:live-mayw2}
\end{figure*}

Table~\ref{tab:live-fourweeks} lists, for each of the four consecutive FutureX weekly rounds we submitted to, the per-level breakdown and final weighted score for our system (FutureGPTv1) alongside the two strongest contemporaneous external submissions (MiroMind's MiroFlow and the H2O AI Super Agent~v1.82). Note that our results were retrieved from the live FutureX platform (\url{https://futurex.live/}) on May 24, 2026. Figures~\ref{fig:live-aprw3}--\ref{fig:live-mayw2} show the corresponding public leaderboard snapshots, cropped to the same rank ranges as the rows reported in Table~\ref{tab:live-fourweeks}. Our system ranks between \#1 and \#6 across the four weeks. The live leaderboard scores are not directly comparable across weeks because each round uses a fresh question pool and has a different mix of resolved levels. In particular, the L3/L4 score of May~W1 is $0.00$ for any listed system at the leaderboard snapshot, so every submission's weighted score is driven only by the L1 and L2 terms, $0.1\,m_1+0.2\,m_2\leq 30$, making the absolute scores in that week much lower than in the other rounds.

\begin{figure}
    \centering
    \includegraphics[width=1\linewidth]{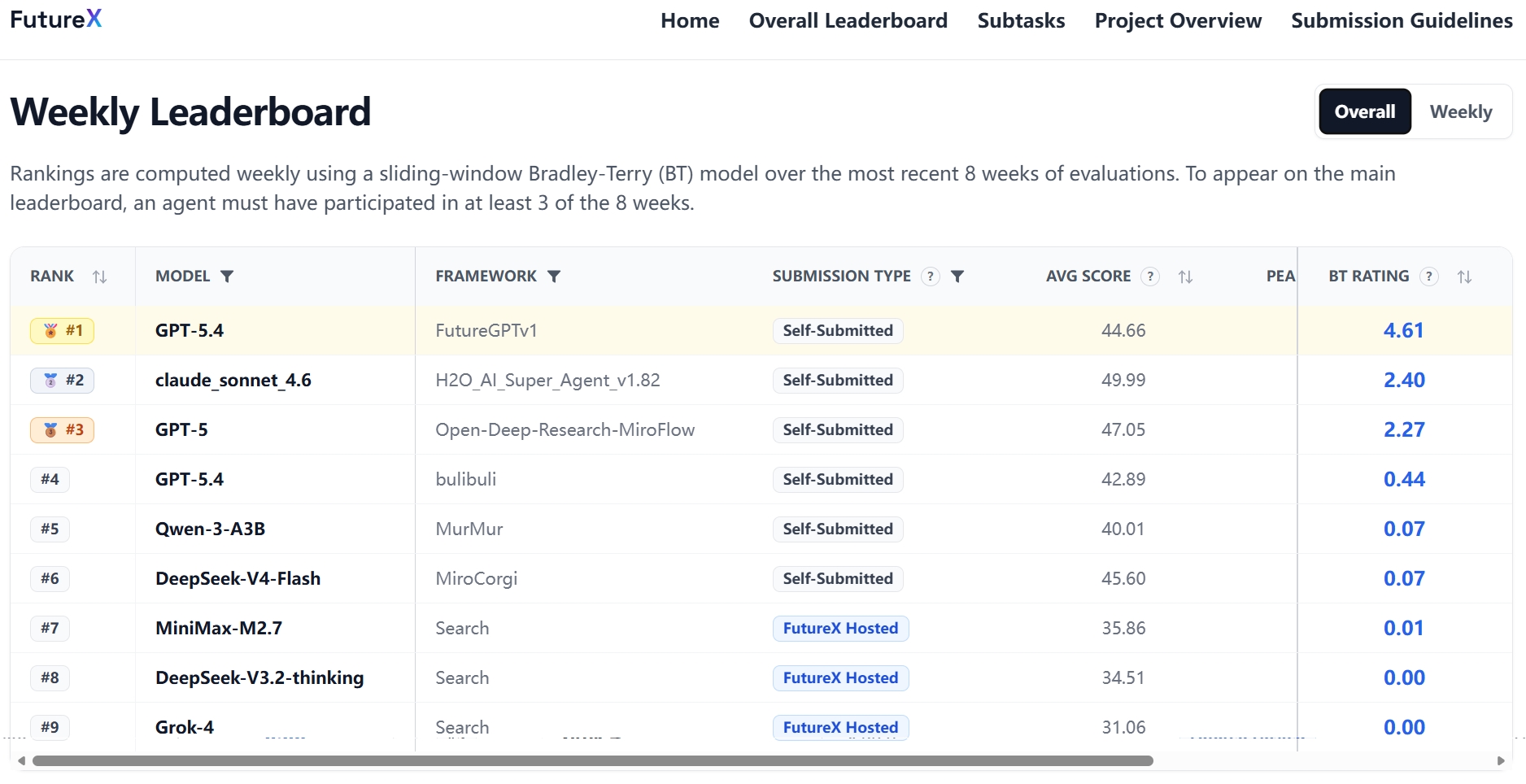}
    \caption{Snapshot of the FutureX overall live leaderboard, accessed on June 17, 2026.}
    \label{fig:live-overall}
\end{figure}

In addition to the week-by-week results, the FutureX public leaderboard also reports an overall ranking computed with a sliding-window Bradley--Terry (BT) model over the most recent eight weeks of evaluations, including only agent frameworks that participated in at least three of the eight weeks. As shown in Figure~\ref{fig:live-overall}, when accessed on June 17, 2026, our system FutureGPTv1 ranked first overall on the official FutureX leaderboard, with a BT rating of $4.61$, ahead of H2O AI Super Agent v1.82 and MiroFlow.

\begin{table*}[h]
  \centering
  \small
  \footnotesize
  \setlength{\tabcolsep}{3pt}
  \begin{tabular}{lllcccccc}
    \toprule
    \textbf{Week (events)} & \textbf{Model} & \textbf{System} & \textbf{Rank} & \textbf{L1} & \textbf{L2} & \textbf{L3} & \textbf{L4} & \textbf{Score} \\
    \midrule
    Apr.~W3 (56)  & Claude-Sonnet-4.6 & H2O AI Super Agent v1.82 & 2  & 89.47 & 50.45 & 50.21 & 52.80 & 55.22  \\
                  & GPT-5.4           & \textbf{Ours} (FutureGPTv1) & \textbf{4}  & 84.21 & 50.30 & 42.64 & 50.98 & 51.67  \\
                  & GPT-5             & MiroFlow                  & 6  & 68.42 & 50.02 & 46.52 & 52.07 & 51.63  \\
    \midrule
    Apr.~W4 (48)  & GPT-5             & MiroFlow                  & 1  & 42.86 & 72.78 & 57.77 & 36.04 & 50.59  \\
                  & GPT-5.4           & \textbf{Ours} (FutureGPTv1) & \textbf{2}  & 42.86 & 69.17 & 54.07 & 32.70 & 47.42  \\
                  & Claude-Sonnet-4.7 & H2O AI Super Agent v1.82  & 4  & 42.86 & 66.06 & 48.88 & 34.74 & 46.06 \\
    \midrule
    May~W1 (48)   & GPT-5.4           & \textbf{Ours} (FutureGPTv1) & \textbf{1}  & 91.67 & 86.18 & 0.00 & 0.00 & 26.40 \\
                  & Claude-Sonnet-4.6 & H2O AI Super Agent v1.82  & 7  & 87.50 & 72.60 & 0.00 & 0.00 & 23.27  \\
                  & GPT-5             & MiroFlow                  & 10 & 79.17 & 75.95 & 0.00 & 0.00 & 23.11  \\
    \midrule
    May~W2 (56)   & Claude-Sonnet-4.6 & H2O AI Super Agent v1.82  & 5  & 100.00 & 43.57 & 41.78 & 38.31 & 46.57  \\
                  & GPT-5.4           & \textbf{Ours} (FutureGPTv1) & \textbf{6}  & 87.50 & 57.14 & 37.78 & 37.12 & 46.36  \\                  & GPT-5.4           & GPT-5.4-SC@3 (GPT@3)             & 7  & 87.50 & 57.14 & 37.77 & 34.45 & 45.29  \\
                  & GPT-5.4           & GPT-5.4 w/ search (GPT-search)         & 8  & 75.00 & 50.00 & 39.55 & 34.45 & 43.15  \\
                  & GPT-5.4           & GPT-5.4-SC@5 (GPT@5)             & 9  & 75.00 & 50.00 & 36.00 & 34.45 & 42.08  \\
                  & Kimi-K2.5         & Kimi-K2.5 (Simple ReAct)        & 17 & 87.50 & 42.14 & 29.24 & 30.82 & 38.28  \\
    \bottomrule
  \end{tabular}
  \caption{Weekly FutureX live leaderboard scores for our submission (FutureGPTv1) and the two strongest systems, accessed at 2026-05-24. For May~W2, we also include four additional baselines: GPT-5.4 with search, GPT-5.4-SC@3, GPT-5.4-SC@5, and Kimi-K2.5 with tools.}
  \label{tab:live-fourweeks}
\end{table*}

\subsection{Complementary Efficiency Results}\label{app:comp_efficiency}
\begin{figure}[t]
    \centering
    \includegraphics[width=0.6\linewidth]{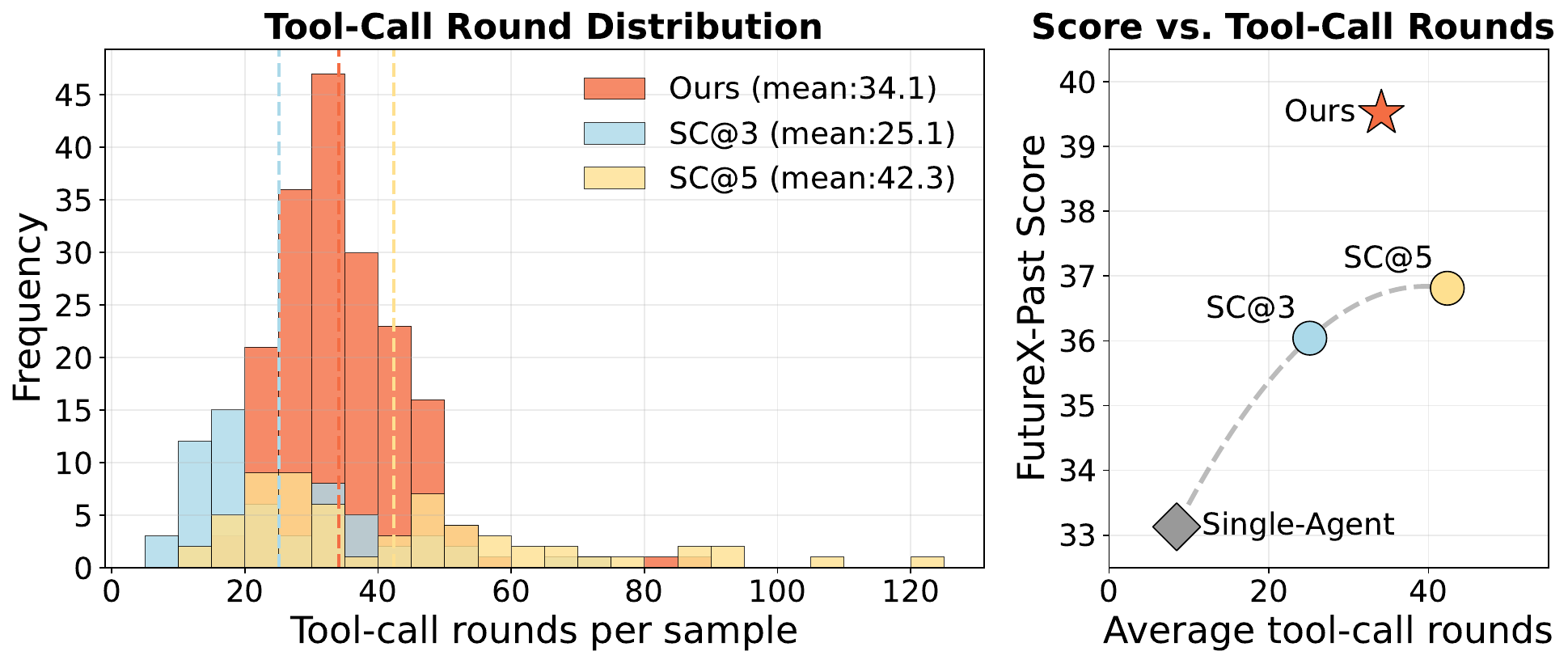}
    \caption{Tool-call efficiency of the multi-agent forecaster compared to self-consistency baselines. \textbf{Left}: per-question tool-call distribution across the 61-question pool; dashed lines mark per-method means (Ours $34.1$, SC@3 $25.1$, SC@5 $42.3$). \textbf{Right}: weighted score versus average tool-call rounds.}
    \label{fig:efficiency_futurex}
\end{figure}

Figure~\ref{fig:efficiency_futurex} reports the per-question tool-call distribution on the left and a score-versus-rounds plot on the right on FutureX-Past dataset. Two features stand out: 1) our framework averages $34.1$ rounds per question, sitting between SC@3 ($25.1$) and SC@5 ($42.3$). Its distribution is also tighter than SC@5, which has a long right tail extending past $90$ rounds. 2) The three baselines line up on a concave efficiency curve with clearly diminishing returns. SC@5 only gains $0.8$ weighted points over SC@3 for about $68\%$ more rounds. Our framework lies clearly above this curve. It beats SC@5 on both axes, with a higher score and fewer rounds. Together, this shows that splitting a question into planner-chosen angles is a more efficient way to spend tool calls than running the same agent many times and voting.

\begin{figure}[t]
    \centering
    \includegraphics[width=0.7\linewidth]{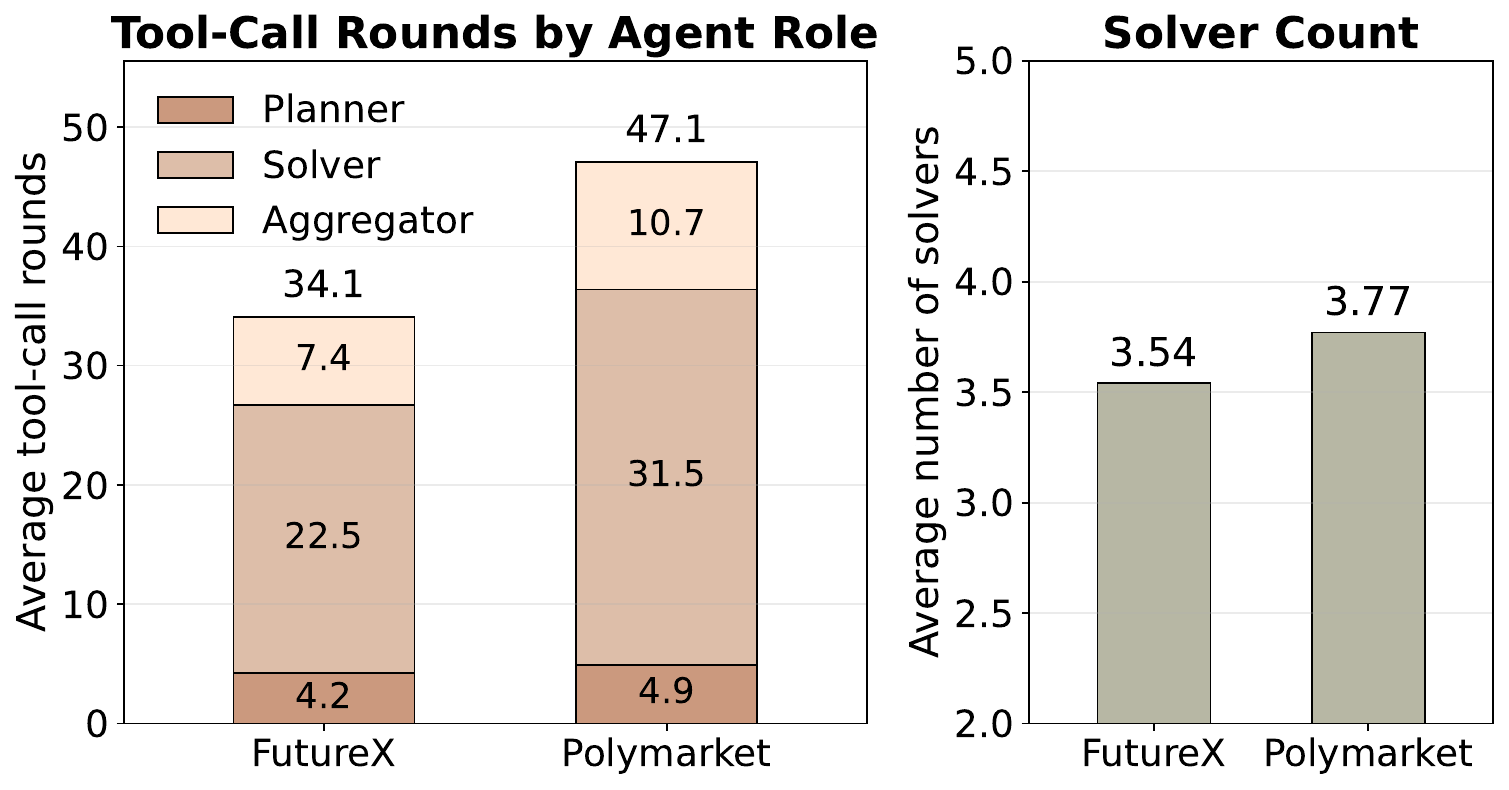}
    \caption{Tool-call cost of our framework on FutureX and Polymarket. \textbf{Left}: average tool-call rounds per question, decomposed into planner, solvers, and aggregator. \textbf{Right}: average number of solvers selected by the planner per question.}
    \label{fig:ours_tools_compare}
\end{figure}
Here, we also break down our framework's tool budget by agent role and compare the two datasets. Figure~\ref{fig:ours_tools_compare} reports the average tool-call rounds spent by the planner, solvers, and aggregator on each question, together with the average number of solvers the planner chooses. On FutureX, a question takes $34.1$ rounds on average, while on Polymarket it takes $47.1$. The planner cost is almost the same on both datasets ($4.2$ vs $4.9$ rounds), so the extra budget on Polymarket comes from the solvers ($22.5$ to $31.5$) and the aggregator ($7.4$ to $10.7$). The planner also tends to spawn slightly more solvers on Polymarket ($3.77$ vs $3.54$).

Polymarket questions are harder for two reasons. First, they are often multi-part markets: a single question can bundle several player props or game outcomes, so each solver has to gather more facts and the aggregator has to cross-check more sub-claims against box scores, lineups, and odds. Second, the relevant information is scattered across many sources such as team news, sportsbook lines, recap articles, and statistics sites, which forces both solvers and the aggregator to issue more search and visit calls. FutureX questions, in contrast, more often center on a single authoritative source such as an official ranking or a published index, which is easier to resolve.

\subsection{Framework with Different Base Models}
Table~\ref{tab:base-model-framework} investigates whether the planner-solver-aggregator design depends on a single backbone. We run the planner-solver-aggregator framework with three base models and compare it with the corresponding single-agent w/ tools baseline from Table~\ref{tab:main}. The framework improves all three backbones on both datasets, suggesting that the benefit comes from the role decomposition itself rather than only from GPT-5.4-medium.

\begin{table*}[h]
  \centering
  \small
  \setlength{\tabcolsep}{5pt}
  \begin{tabular}{lccc|ccc}
    \toprule
    \textbf{Base model} &
    \multicolumn{3}{c|}{\textbf{FutureX-Past}} &
    \multicolumn{3}{c}{\textbf{Polymarket}} \\
    & \textbf{Single agent} & \textbf{Framework} & $\Delta$ &
      \textbf{Single agent} & \textbf{Framework} & $\Delta$ \\
    \midrule
    GPT-5.4-medium       & 33.13 & $39.51 \pm 1.34$ & +6.38  & 58.63 & $64.99 \pm 4.15$ & +6.36 \\
    Claude-4.6-Sonnet    & 22.80 & $35.36 \pm 2.64$ & +12.56 & 48.21 & $59.03 \pm 0.66$ & +10.82 \\
    DeepSeek-V4-Flash    & 35.02 & $36.90 \pm 2.27$ & +1.88  & 48.12 & $54.48 \pm 2.73$ & +6.36 \\
    \bottomrule
  \end{tabular}
  \caption{Planner-solver-aggregator framework with different base models. Single-agent scores are the corresponding w/ tools rows in Table~\ref{tab:main}. Framework scores report mean $\pm$ standard deviation over three independent runs. $\Delta$ is the framework mean minus the single-agent score. All scores are scaled by $100$.}
  \label{tab:base-model-framework}
\end{table*}

\subsection{Case Studies for Planner-Solver-Aggregator Framework}
\label{app:mas-cases}

We analyze two trajectories through our multi-agent framework on FutureX-Past. In both cases, the final answer is not directly available from the agent's tool calls. The correct prediction therefore requires reasoning over multiple sources, rather than simple retrieval. Table~\ref{tab:case-sp500} and Table~\ref{tab:case-gistemp} summarize the outputs of the planner, solvers, and aggregator for each run.

\paragraph{Case A: Q1 2026 S\&P 500 quarter return (L2, qid \texttt{69a2e39e5692ef005cdbf2d8}).} 
The question presents nine non-overlapping return ranges for the S\&P 500 price return on March 31, 2026, where choice A represents a return below $0\%$. Because the Time Machine blocks the final closing value, the agent must deduce the correct range from earlier market data. Table~\ref{tab:case-sp500} outlines the actions of each role. All three solvers independently select choice A using different analytical paths, and the aggregator confirms the underlying data before final alignment.

\begin{table*}[h]
  \centering
  \small
  \renewcommand{\arraystretch}{1.25}
  \setlength{\tabcolsep}{4pt}
  \begin{tabular}{@{}p{2.6cm} p{11cm} p{0.7cm} p{0.7cm}@{}}
    \toprule
    \textbf{Role} & \textbf{Key action / finding} & \textbf{Pick} & \textbf{Conf.} \\
    \midrule
    Planner & Decomposes the task into three angles, namely a near-resolution market-state tracker, a historical Q1 distribution analyst, and a late-quarter macro / event-risk analyst. & --
 & -- \\
    \midrule
    Solver 1 \par (market path) & Anchors Dec.~31 2025 close ($6{,}845.50$) and Mar.~30 2026 close ($6{,}343.72$); computes quarter-to-date return of $-7.33\%$; notes that escaping bucket A on the final session would require a $+7.91\%$ one-day rally. & A & 0.94 \\
    Solver 2 \par (history) & Independent prior shortlist \{A,\,E,\,F\} on quarter-distribution grounds; verifies the same Dec.~31 and Mar.~30 anchors via Yahoo Finance snippets; concludes that the $-7.33\%$ gap is far outside ordinary quarter-end churn. & A & 0.92 \\
    Solver 3 \par (macro) & Iran/Middle-East oil shock pushed oil sharply higher, Fed easing expectations were repriced lower because of inflation risk, and tariff/trade uncertainty remained a headwind; all unfavorable for a late rebound. & A & 0.95 \\
    \midrule
    Aggregator & Cross-confirms the two anchor closes from independent search snippets (CNBC, Morningstar/Dow Jones, Yahoo for Mar.~30; Yahoo and syndicated close reports for Dec.~31); reproduces the same $+7.91\%$ rally threshold and finds no event capable of producing one. & \textbf{A} & 0.94 \\
    \bottomrule
  \end{tabular}
  \caption{Multi-agent trajectory for the Q1 2026 S\&P 500 question (qid \texttt{69a2e39e5692ef005cdbf2d8}, with ground truth A). All three solvers independently reach A via complementary analytical paths; the aggregator re-verifies the anchors and concurs.}
  \label{tab:case-sp500}
\end{table*}

\paragraph{Case B: NASA GISTEMP December 2025 anomaly (L2, qid \texttt{6941510e41b9d1005effa734}).} The question is about the December 2025 global land-ocean temperature anomaly relative to the 1951 to 1980 baseline, where choice A contains values below $1.095^\circ\mathrm{C}$. The official NASA GISTEMP data files are completely unreadable through the visit tool, forcing the agents to infer the answer from alternative climate datasets instead. Table~\ref{tab:case-gistemp} details the actions of each role. Unlike the uniform consensus in the previous case, the three solvers disagree completely. Two solvers rely on historical intuition that the period was exceptionally warm, whereas the aggregator performs cross-dataset calculations to override these incorrect estimates. This specific conflict indicates why vanilla voting fails on complex forecasting tasks: a simple majority vote over the three conflicting answers results in a tie and selects the correct choice only by chance. In contrast, our framework reaches the correct answer because the aggregator independently verifies the cross-source evidence.

\begin{table*}[h]
  \centering
  \small
  \renewcommand{\arraystretch}{1.25}
  \setlength{\tabcolsep}{4pt}
  \begin{tabular}{@{}p{2.6cm} p{11cm} p{0.7cm} p{0.7cm}@{}}
    \toprule
    \textbf{Role} & \textbf{Key action / finding} & \textbf{Pick} & \textbf{Conf.} \\
    \midrule
    Planner & Decomposes the task into three angles, namely an official NASA release reader, a cross-dataset triangulator using NOAA and Copernicus, and a revision-risk watchdog. & --
 & --
 \\
    \midrule
    Solver 1 \par (NASA reader) & NASA tables and CSVs not retrievable through visit; reads the indirect 'fifth-warmest December' framing as a moderately-warm signal and places mass on bin D without baseline conversion. & D & 0.47 \\
    Solver 2 \par (cross-dataset) & NOAA December 2025 anomaly $1.05^\circ\mathrm{C}$ above 20th-century mean and Copernicus December anomaly $0.49^\circ\mathrm{C}$ above 1991--2020 (both ranked fifth-warmest December); mapping to NASA's 1951--1980 baseline gives $\approx 1.0$--$1.1^\circ\mathrm{C}$, just below the A/B boundary. & A & 0.57 \\
    Solver 3 \par (revision) & Initial prior anchored on broad 'late-2025 was hot' regime intuition, giving an E/F/G shortlist before evidence; NASA revision-cadence snippets address only update timing, not the central value. & F & 0.43 \\
    \midrule
    Aggregator & Verifies NOAA and Copernicus exact-month values; notes the annual NASA--NOAA gap is only about $+0.02^\circ\mathrm{C}$, not enough to lift the December value above the $1.095^\circ\mathrm{C}$ threshold; downweights Solver~1 (only an ordinal rank, no numeric anomaly to compare against this threshold) and Solver~3 (not month-specific). & \textbf{A} & 0.58 \\
    \bottomrule
  \end{tabular}
  \caption{Multi-agent trajectory for the NASA GISTEMP December 2025 question (qid \texttt{6941510e41b9d1005effa734}, with ground truth A). The three solvers split $1$--$1$--$1$ across D, A, F; the aggregator's own cross-dataset arithmetic overrides the two warm-prior angles and selects A.}
  \label{tab:case-gistemp}
\end{table*}

\end{document}